\documentclass{article}
\usepackage[final]{neurips_2021}

\usepackage[utf8]{inputenc} 
\usepackage[T1]{fontenc}    
\usepackage[unicode]{hyperref}       
\usepackage{url}            
\usepackage{booktabs}       
\usepackage{amsfonts}       
\usepackage{nicefrac}       
\usepackage{microtype}      
\usepackage[dvipsnames]{xcolor}         

\usepackage{amsmath}
\usepackage{amsthm}
\usepackage{dsfont}
\usepackage{mathtools}
\usepackage{pgf}
\usepackage{chngpage}
\usepackage{subcaption}
\usepackage[capitalise, noabbrev]{cleveref}
\usepackage[ruled,linesnumbered]{algorithm2e}
\usepackage{accents}
\usepackage{adjustbox}

\usepackage{enumitem}
\usepackage{etoolbox}
\appto\normalsize{\belowdisplayskip=\belowdisplayshortskip}
\appto\small{\belowdisplayskip=\belowdisplayshortskip}
\newcommand*\samethanks[1][\value{footnote}]{\footnotemark[#1]}

\usepackage[textwidth=30mm]{todonotes}

\author{%
  Flore Sentenac\thanks{Equal contributions} \\
  CREST, ENSAE Paris, Palaiseau, France\\
  \texttt{flore.sentenac@gmail.com} \\
  \And
  Etienne Boursier\samethanks\\
  Centre Borelli, ENS Paris-Saclay, France\\ 
  \texttt{etienne.boursier1@gmail.com} \\
  \And
  Vianney Perchet\\
  CREST, ENSAE Paris,  Palaiseau, France\\
  CRITEO AI Lab, Paris, France\\
  \texttt{vianney.perchet@normalesup.org} \\
}

\newtheorem{assumption}{Assumption}
\newtheorem{remark}{Remark}
\newtheorem{theorem}{Theorem}
\newtheorem{lemma}{Lemma}
\newtheorem{definition}{Definition}
\newtheorem{proposition}{Proposition}

\Crefname{algocf}{Algorithm}{Algorithms}
\crefname{assumption}{Assumption}{Assumptions}

\crefalias{AlgoLine}{line}
\makeatletter
\let\cref@old@stepcounter\stepcounter
\def\stepcounter#1{%
  \cref@old@stepcounter{#1}%
  \cref@constructprefix{#1}{\cref@result}%
  \@ifundefined{cref@#1@alias}%
    {\def\@tempa{#1}}%
    {\def\@tempa{\csname cref@#1@alias\endcsname}}%
  \protected@edef\cref@currentlabel{%
    [\@tempa][\arabic{#1}][\cref@result]%
    \csname p@#1\endcsname\csname the#1\endcsname}}
\makeatother

\newcommand{\ie}{i.e.,\ }
\newcommand{\eg}{e.g.,\ }

\newcommand{\bigO}[1]{\mathcal{O}\left(#1\right)}

\newcommand{\debug}[1]{#1}
\newcommand\munderbar[1]{%
  \underaccent{\bar}{#1}}

\newcommand{\one}[1]{\mathds{1} \left( #1 \right)}
\newcommand{\supp}[1]{\mathrm{supp}( #1 )}
\newcommand{\df}{\mathrm{d}}
\newcommand{\bernoulli}{\mathrm{Bernoulli}}
\newcommand{\gE}[1][t]{\mathcal{E}_{#1}}
\newcommand{\permuset}[1][K]{\mathfrak{S}_{#1}}

\newcommand{\proj}{\Pi}
\newcommand{\vol}[1]{\mathrm{vol}\left( #1 \right)}
\newcommand{\probaset}[1]{\mathcal{P}\left( #1 \right)}
\newcommand{\smallo}[1]{o\left( #1 \right)}
\renewcommand{\mod}[1]{\ (\mathrm{mod} \ #1)}

\DeclareMathOperator*{\argmin}{arg\,min}
\newcommand{\E}{\mathbb{E}}
\newcommand{\Pb}{\mathbb{P}}
\newcommand{\bisto}{\mathfrak{B}_K}

\newcommand{\explore}{\debug \textsc{Explore}}
\newcommand{\algoname}{\debug \textsc{ADeQuA}}
\newcommand{\birkhoff}{\debug \textsc{Ordered Birkhoff}}
\newcommand{\hungarian}{\debug \textsc{Hungarian}}

\newcommand{\approxproj}[2]{\hat{\proj}_{#2}\left( #1 \right) }
\newcommand{\mappsicomplete}{\debug Birkhoff von Neumann decomposition}
\newcommand{\mappsi}{\debug BvN decomposition}
\newcommand{\mapphi}{\debug dominant mapping}

\SetCommentSty{mycommfont}
\title{Decentralized Learning in Online Queuing Systems}
\begin{document}
\maketitle

\begin{abstract}
    Motivated by packet routing in computer networks and resource allocation in radio networks, online queuing systems are composed of queues receiving packets at different rates. Repeatedly, they send packets to servers, each of them treating only at most one packet at a time. In the centralized case, the number of accumulated packets remains bounded (i.e., the system is \textit{stable}) as long as the ratio between service rates and arrival rates is larger than $1$. In the decentralized case, individual no-regret strategies ensures stability when this ratio is larger than $2$. Yet, myopically minimizing  regret disregards the long term effects due to the carryover of packets to further rounds. On the other hand, minimizing long term costs leads to stable Nash equilibria as soon as the ratio exceeds $\frac{e}{e-1}$. Stability with decentralized learning strategies  with a ratio below $2$ was a major remaining question.
    We first argue that for ratios up to $2$, cooperation is required for stability of learning strategies, as  selfish minimization of policy regret, a \textit{patient} notion of regret, might indeed still be unstable in this case.
    We therefore consider cooperative queues and propose the first learning decentralized algorithm guaranteeing stability of the system as long as the ratio of rates is larger than $1$, thus reaching performances comparable to centralized strategies.
\end{abstract}
\linepenalty=1000
\section{Introduction}
Inefficient decisions in repeated games can stem from both strategic and learning considerations.
First, strategic agents selfishly maximize their own individual reward at others' expense. The \textit{price of anarchy} \citep{koutsoupias1999worst} measures this inefficiency as the social welfare ratio between the best possible situation and the worst  Nash equilibrium. 
Although reaching the best collective outcome might be illusory for selfish agents, considering the worst Nash equilibrium might be too pessimistic. 
In games with external factors, more complex interactions  intervene and might lead the agents to the best equilibrium.
Instead, the \textit{price of stability} \citep{schulz2003performance} measures the inefficiency by the social welfare ratio between the best possible situation and the best  Nash equilibrium.

Second, the agents also have to learn their environment, by repeatedly experimenting different outcomes.
Learning equilibria in repeated games is at the core of many problems in computer science and economics \citep{fudenberg1998theory,cesa2006prediction}. The interaction between multiple agents can indeed interfere in the learning process, potentially converging to no or bad equilibria.
It is yet known that in repeated games, if all agents follow no internal regret strategies, their actions converge in average to the set of correlated equilibria \citep{hart2000simple,blum2007learning,survey}. 

Many related results are known in the classical repeated games \citep[see \eg][]{cesa2006prediction,roughgarden2010algorithmic}, where a single game is repeated over independent rounds (but the agents strategies might evolve and depend on the history). 
Motivated by packet routing in computer networks, \citet{gaitonde2020stability} introduced a repeated game with a \textit{carryover} feature: the outcome of a round does not only depend on the actions of the agents, but also on the previous rounds.
They consider heterogeneous queues sending packets to servers. If several queues simultaneously send packets to the same server, only the oldest packet is treated by the server. 

Because of this carryover effect, little is known about this type of game. In a first paper, \citet{gaitonde2020stability} proved that if queues follow suitable no-regret strategies, a ratio of $2$ between server and arrival rates leads to stability of the system, meaning that the number of packets accumulated by each queue remains bounded. 
However, the assumption of regret minimization sort of reflects a myopic behavior and is not adapted to games with carryover. 
\citet{gaitonde2020virtues} subsequently consider a patient game, where queues instead minimize their asymptotic number of accumulated packets.
A ratio only larger than $\frac{e}{e-1}$ then guarantees the stability of the system, while a smaller ratio leads to inefficient Nash equilibria. As a consequence, going below the $\frac{e}{e-1}$ factor requires some level of cooperation between the queues.
This result actually holds with perfect knowledge of the problem parameters and it remained even unknown whether decentralized learning strategies can be stable with a ratio below $2$. 

We first argue that decentralized queues need some level of cooperation to ensure stability with a ratio of rates below $2$. Policy regret can indeed be seen as a patient alternative to the regret notion. Yet even minimizing the policy regret might lead to instability when this ratio is below~$2$.
An explicit decentralized cooperative algorithm called \algoname\ (\textsc{A} \textsc{De}centralized \textsc{Qu}euing \textsc{A}lgorithm) is thus proposed. It is the first decentralized learning algorithm guaranteeing stability when this ratio is only larger than $1$.
\algoname\ does not require communication between the queues, but uses synchronisation between them to accurately estimate the problem parameters and avoid interference when sending packets. Our main result is given by \cref{thm:stabilityintro} below, whose formal version, \cref{thm:stability} in \cref{sec:algo}, also provides bounds on the number of accumulated packets.
\begin{theorem}[\cref{thm:stability}, informal]\label{thm:stabilityintro}
If the ratio between server rates and arrival rates is larger than~$1$ and all queues follow \algoname, the system is strongly stable.
\end{theorem}
The remaining of the paper is organised as follows. The model and existing results are recalled in \cref{sec:model}. \cref{sec:cooperative} argues that cooperation is required to guarantee stability of learning strategies when the ratio of rates is below $2$. \algoname\ is then presented in \cref{sec:algo}, along with insights for the proof of \cref{thm:stabilityintro}.
\cref{sec:simulations} finally compares the behavior of \algoname\ with no-regret strategies on toy examples and empirically confirms the different known theoretical results.

\subsection{Additional related work}

Queuing theory includes applications in diverse areas such as computer science, engineering, operation research \citep{shortle2018fundamentals}. \citet{borodin1996adversarial} for example use the stability theorem of \citet{pemantle1999moment}, which was also used by \citet{gaitonde2020virtues}, to study the problem of packet routing through a network.
%
Our setting is the single-hop particular instance of throughput maximization in wireless networks. Motivated by resource allocation in multihop radio problem, packets can be sent through more general routing paths in the original problem. \citet{tassiulas1990stability} proposed a first stable centralized algorithm, when the service rates are known \textit{a priori}. Stable decentralized algorithms were later introduced in specific cases \citep{neely2008fairness,jiang2009distributed,shah2012randomized}, when the rewards $X_k(t)$ are observed before deciding which server to send the packet. The main challenge then is coordination and queues aim at avoiding collisions with each other. The proposed algorithms are thus not adapted to our setting, where both coordination between queues and learning the service rates are required. We refer the reader to \citep{georgiadis2006resource} for an extended survey on resource allocation in wireless networks.

\medskip

\citet{NIPS2016_430c3626} first considered online learning for such queuing systems model, in the simple case of a single queue. It is a particular instance of stochastic multi-armed bandits, a celebrated online learning model, where the agent repeatedly takes an action within a finite set and observes its associated reward.
This model becomes intricate when considering multiple queues, as they interfere when choosing the same server. It is then related to the multiplayer bandits problem which considers multiple players simultaneously pulling arms. When several of them pull the same arm, a \textit{collision} occurs and they receive no reward \citep{anandkumar2010opportunistic}.

The collision model is here different as one of the players still gets a reward. It is thus even more closely related to competing bandits \citep{liu2020competing,liu2020bandit}, where arms have preferences over the players and only the most preferred player pulling the arm actually gets the reward. Arm preferences are here not fixed and instead depend on the packets' ages.
While collisions can be used as communication tools between players in multiplayer bandits \citep{bistritz2018distributed,boursier2019sic, mehrabian2020practical, wang2020optimal}, this becomes harder with an asymmetric collision model as in competing bandits. However, some level of communication remains possible \citep{sankararaman2020dominate,basu2021beyond}. 
In queuing systems, collisions are not only asymmetric, but depend on the age of the sent packets, making such solutions unsuited. 

While multiplayer bandits literature considers cooperative players, \citet{boursier2020selfish} showed that cooperative algorithms could be made robust to selfish players. On the other hand, competing bandits consider strategic players and arms as the goal is to reach a bipartite stable matching between them.
Despite being cooperative, \algoname\ also has strategic considerations as the queues' strategy converges to a correlated equilibrium of the patient game described in \cref{sec:model}.

An additional difficulty here appears as queues are asynchronous: they are not active at each round, but only when having packets left. This is different from the classical notion of asynchronicity \citep{bonnefoi2017multi}, where players are active at each round with some fixed probability. 
Most strategies in multiplayer bandits rely on synchronisation between the players \citep{boursier2019sic} to avoid collisions. While such a level of synchronisation is not possible here, some lower level of synchronisation is still used to avoid collisions between queues.

\section{Queuing Model} \label{sec:model}

We consider a queuing system composed of $N$ queues and $K$ servers, associated with vectors of arrival and service rates $\boldsymbol{\lambda},\boldsymbol{\mu}$, where at each time step $t=1,2, \ldots,$ the following happens:
\begin{itemize}[leftmargin=1cm,topsep=0pt] \setlength\itemsep{0.5em}
    \item each queue $i\in[N]$ receives a new packet with probability $\lambda_i \in [0,1]$, that is marked with the timestamp of its arrival time. If the queue currently has  packet(s) on hold, it sends one of them to a chosen server $j$ based on its past observations. 
    \item Each server $j\in[K]$  attempts to clear the oldest packet it has received, breaking ties uniformly at random. It succeeds with probability $\mu_j\in [0,1]$ and otherwise sends it back to its original queue, as well as all other unprocessed packets. 
\end{itemize}

At each time step, a queue only observes whether or not the packet sent (if any) is cleared by the server.
We note $Q_t^{i}$ the number of packets in queue $i$ at time $t$. Given a packet-sending dynamics, the system is \textbf{stable} if, for each $i$ in $[N]$, $Q_t^{i}/t$ converges to 0 almost surely. It is \textbf{strongly stable}, if for any $r,t \geq 0$ and $i \in [N]$, $\E[(Q_t^{i})^r]\leq C_r$, where $C_r$ is an arbitrarily large constant, depending on $r$ but not $t$. Without ambiguity, we also say the policy or the queues are (strongly) stable. Naturally, a strongly stable system is also stable \citep{gaitonde2020stability}.

In the following, $x_{(i)}$ will denote the $i$-th order statistics of a vector $\boldsymbol{x}$, \ie $\lambda_{(1)}\geq\lambda_{(2)}\geq\ldots\geq \lambda_{(N)}$ and ${\mu_{(1)}\geq\ldots\geq \mu_{(K)}}$. Without loss of generality, we assume $K\geq N$ (otherwise, we simply add fictitious servers with $0$ service rate). The key quantity of a system is its  \textbf{slack}, defined  as the largest real number $\eta$ such that:
\begin{equation*}\label{EQ:slack}
\sum_{i=1}^{k} \mu_{(i)}\geq\eta \sum_{i=1}^{k} \lambda_{(i)}, \ \forall\  k \leq N.
\end{equation*}
We also denote by $\probaset{[K]}$ the set of probability distributions on $[K]$ and by  $\Delta$ the \textbf{margin} of the system defined by
\begin{equation}\label{EQ:margin}
\Delta \coloneqq \min_{k\in[N]} \frac{1}{k}\sum_{i=1}^k (\mu_{(i)} - \lambda_{(i)}).
\end{equation}
Notice that the alternative system where  $\tilde{\lambda}_i= \lambda_i+\Delta$ and $\tilde{\mu}_k=\mu_k$ has a slack $1$. In that sense, $\Delta$ is the largest \textit{margin} between service and arrival rates that all queues can individually get in the system. Note that if $\eta>1$, then $\Delta>0$. 
%
%
We now recall existing results for this problem,  summarized in \cref{fig:frise} below.

\begin{theorem}[\citealt{marshall1979inequalities}]
 \label{lem:centralstable} For any instance, there exists a strongly stable centralized policy if and only if $\eta>1$.
\end{theorem}

\begin{theorem}[\citealt{gaitonde2020stability}, informal] \label{lem:regret2unstable} If $\eta>2$, queues following appropriate no regret strategies are strongly stable. \\
For each $N>0$, there exists a system and a dynamic s.t. $\eta>2-o(1/N)$, all queues follow appropriate no-regret strategies, but they are not strongly stable.
\end{theorem}
In the above theorem, an \textit{appropriate no regret strategy} is a strategy such that there exists a partitioning of the time into successive windows, for which the incurred regret is $\smallo{w}$ with high probability on any window of length $w$. This for example includes the EXP3.P.1 algorithm \citep{auer2002nonstochastic} where the $k$-th window has length $2^k$.

The patient queuing game $\mathcal{G}=([N],(c_i)_{i=1}^n, \boldsymbol{\mu}, \boldsymbol{\lambda})$ is defined as follows.  The strategy space for each queue is $\probaset{[K]}$. Let $\boldsymbol{p}_{-i}\in (\probaset{[K]})^{N-1} $ denote the vector of fixed distributions for all queues over servers, except for queue $i$. The cost function for queue $i$ is defined as:
\begin{equation*}
    c_i(p_i, \boldsymbol{p}_{-i})= \lim_{t\rightarrow + \infty} T_t^i/t,
\end{equation*}
where $T_t^i$ is the age of the oldest packet in queue $i$ at time $t$. Bounding $T_t^i$ is equivalent to bounding~$Q_t^i$.

\begin{theorem}[\citealt{gaitonde2020virtues}, informal]\label{lem:Nashstable} If $\eta>\frac{e}{e-1}$, any Nash equilibrium of the patient game $\mathcal{G}$ is stable.
\end{theorem}
\definecolor{grey}{rgb}{0.4,0.4,0.4}
\definecolor{green}{rgb}{0,0.4,0}
\definecolor{blue}{rgb}{0,0,1}
\definecolor{red}{rgb}{1,0,0}
\definecolor{purple}{rgb}{0.5,0.,0.9}
\definecolor{greydark}{rgb}{0.5,0.5,0.5}
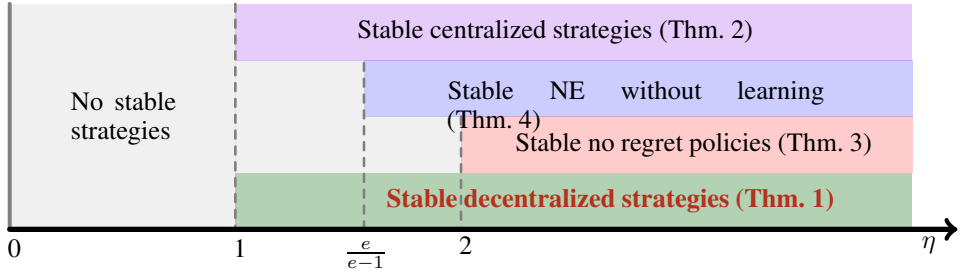
\begin{figure}[ht]
\centering
\begin{tikzpicture}[line cap=round,line join=round,scale=3]
\fill[line width=2pt,color=grey,fill=grey,fill opacity=0.1] (0,1) -- (1,1) -- (1,0) -- (0,0) -- cycle;
\fill[line width=2pt,color=grey,fill=grey,fill opacity=0.1] (1,0.75) -- (1.57,0.75) -- (1.57,0.25) -- (1.,0.25) -- cycle;
\fill[line width=2pt,color=grey,fill=grey,fill opacity=0.1] (1.57,0.5) -- (2,0.5) -- (2,0.25) -- (1.57,0.25) -- cycle;
\fill[line width=2pt,color=purple,fill=purple,fill opacity=0.2] (1,1) -- (4,1) -- (4,0.75) -- (1,0.75) -- cycle;
\fill[line width=2pt,color=blue,fill=blue,fill opacity=0.2] (1.57,0.75) -- (1.57,0.5) -- (4,0.5) -- (4,0.75) -- cycle;
\fill[line width=2pt,color=red,fill=red,fill opacity=0.2] (2.,0.5) -- (2.,0.25) -- (4,0.25) -- (4,0.5) -- cycle;
\fill[line width=2pt,color=green,fill=green,fill opacity=0.3] (4,0.25) -- (4,0) -- (1,0) -- (1,0.25) -- cycle;
\draw (4.0,0.01) node[anchor=north west] {\parbox{1.19 cm}{  $\eta$}};
\draw [line width=1pt,dash pattern=on 3pt off 3pt,color=greydark] (1.57,0)-- (1.57,0.75);
\draw [line width=1pt,dash pattern=on 3pt off 3pt,color=greydark] (2.,0.5)-- (2.,0);
\draw [line width=1pt,dash pattern=on 3pt off 3pt,color=greydark] (1,1)-- (1,0);
\draw [line width=1.5pt,color=greydark] (0,1)-- (0,0);
\draw (0.23,0.65) node[anchor=north west] {\parbox{1.4 cm}{No stable  strategies}};
\draw (1.5,0.97) node[anchor=north west] {Stable centralized strategies (Thm.~\ref{lem:centralstable})};
\draw (2.2,0.47) node[anchor=north west] {Stable no regret policies (Thm.~\ref{lem:regret2unstable})};
\draw (1.9,0.7) node[anchor=north west] {\parbox{5 cm}{Stable NE without learning (Thm.~\ref{lem:Nashstable})}};
\draw (1.63,0.22) node[anchor=north west] {\textbf{\color{BrickRed} Stable decentralized strategies (Thm.~\ref{thm:stabilityintro})}};
\draw (-0.05,-0.0) node[anchor=north west] {0};
\draw (1-0.05,-0.0) node[anchor=north west] {1};
\draw (1.57-0.13,-0.01) node[anchor=north west] {\parbox{1.466 cm}{$\frac{e}{e-1}$}};
\draw (2-0.05,0.01) node[anchor=north west] {2};
\draw [->,line width=2pt] (0,0) -- (4.2,0);
\end{tikzpicture}
\caption{Existing results depending on the slack $\eta$. Our result is highlighted in red.}
\label{fig:frise}
\end{figure}

\section{The case for a cooperative algorithm}\label{sec:cooperative}

According to \cref{lem:regret2unstable,lem:Nashstable}, queues  that are patient enough and select a fixed randomization over the servers are stable over a larger range of slack $\eta$ than queues optimizing their individual regret. A key difference between the two settings is that when minimizing their regret, queues are myopic, which is formalized as follows. Let $a_{1:t}=(a_1,...,a_t)$  be the vector of actions played by a queue up to time~$t$
and let $\nu_t(a_{1:t})$ be the indicator that it cleared a packet at iteration $t$. Classical (external) regret of queue $i$ over horizon $T$ is then defined as:
\begin{equation*}
R_i^{\text{ext}}(T) := \max_{p\in\probaset{[K]}} \sum_{t=1}^T\mathbb{E}_{\tilde{a}_{t}\sim p}[\nu_t(a_{1:t-1},\tilde{a}_{t} )]-\sum_{t=1}^T\nu_t(a_{1:t}).
\end{equation*}
Thus minimizing the external regret is equivalent to  maximizing the instant rewards at each iteration, ignoring the consequences of the played action on the state of the system. 
However, in the context of queuing systems, the actions played by the queues change the state of the system. Notably, letting other queues clear packets can be in the best interest of a queue, as it may give it priority in the subsequent iterations where it holds older packets. Since the objective is to maximize the total number of packets cleared, it seems adapted to minimize a \textit{patient} version of the regret, namely the policy regret \citep{arora2012online}, rather than the external regret, which is defined by
\begin{equation*}
R_i^{\text{pol}}(T) := \max_{p\in\probaset{[K]}} \sum_{t=1}^T\mathbb{E}_{\tilde{a}_{1:t}\sim \otimes_{i=1}^t p}[\nu_t(\tilde{a}_{1:t})]-\sum_{t=1}^T\nu_t(a_{1:t}).
\end{equation*}
That is, $R_i^{\text{pol}}(T)$ is the expected difference between the number of packets queue $i$ cleared and the number of packets it would have cleared over the whole period by playing a fixed (possibly random) action, taking into account how this change of policy would affect the state of the system. 

However, as stated in \cref{prop:counterexample}, optimizing this patient version of the regret rather than the myopic one could not guarantee stability on a wider range of slack value. This suggests that adding only patience to the learning strategy of the queues is not enough to go beyond a slack of $2$, and that any strategy beating that factor $2$ must somewhat include synchronisation between the queues. 

\begin{proposition}\label{prop:counterexample}
Consider the partition of the time $t=1,2, \ldots $ into successive windows, where  $w_k =k^2$ is the length of the $k$-th one. For any $N \geq 2$, there exists an instance with $2N$ queues and servers,  with slack $\eta=2-\bigO{\frac{1}{N}}$, s.t., almost surely, each queue's policy regret is $\smallo{w_k}$ on all but finitely many of the windows, but the system is not strongly stable.
\end{proposition}

To ease comparison, the formulation in the above proposition matches that of the counter-example used to prove \cref{lem:regret2unstable}  (\citealt{gaitonde2020stability}). In that counter-example, a set of system parameters, for which \textit{any} no external regret policies were unstable, was exhibited. Whereas we exhibit in our case a \textit{specific} strategy that satisfies the no policy regret condition, but is unstable.
\begin{proof}[Sketch of proof]
Consider a system with $2N$ queues and servers with $\lambda_i=1/2N$ and $\mu_i=1/N-1/4N^2$ for all $i\in[2N]$. The considered strategy profile is the following. For each $k\geq0$, the $k^{\text{th}}$ time window is split into two stages. During the first stage, of length $\lceil \alpha w_k\rceil$, queues $2i$ and $2i+1$ both play server $2i+t\mod{2N}$ at iteration $t$, for all $i \in [N]$. During the second stage of the time window, queue $i$ plays server $i+t\mod{2N}$ at iteration $t$.
This counter example, albeit very specific, illustrates well how when the queues are highly synchronised, it is better to remain synchronized rather than deviate, even if the synchronisation is suboptimal in terms of stability. The complete proof is provided in \cref{app:counterexample}. 

Queues following this strategy accumulate packets during the first stage, and clear more packets than they receive during the second stage. The value of $\alpha$ is tuned so that the queues still accumulate a linear portion of packets during each time window. For those appropriate $\alpha$, the system is unstable.

Now suppose  that queue $i$ deviates from the strategy and plays a fixed action $p \in \probaset{[K]}$. In the first stage of each time window, queue $i$ can clear a bit more packets than it would by not deviating. However, during the second stage, it is no longer synchronised with the other queues and collides with them a large number of times. Because of those collisions, it will accumulate many packets. In the detailed analysis, we demonstrate that, in the end,  for appropriate values of $\alpha$, queue $i$ accumulates more packets than it would have without deviating.  
\end{proof}

According to \cref{lem:Nashstable}, the factor $\frac{e}{e-1}$ can be seen as the price of anarchy of the problem, as for slacks below, the worst Nash equilibria might be unstable. On the other hand, it is known that for any slack above $1$, there exists a centralized stable strategy. This centralized strategy actually consists in queues playing the same joint probability at each time step, independently from the number of accumulated packets. As a consequence, it is also a correlated equilibrium of the patient game and~$1$ can be seen as the correlated price of stability.
%

\section{A decentralized algorithm}\label{sec:algo}

This section describes the decentralized algorithm \algoname, whose pseudocode is given in \cref{algo1}. Due to space constraints, all the proofs are postponed to \cref{app:proofs}.
\algoname\ assumes all queues \textit{a priori} know the number $N$ of queues in the game and have a unique rank or \textit{id} in $[N]$. Moreover, the existence of a shared randomness between all queues is assumed.
The \textit{id} assumption is required to break the symmetry between queues and is classical in multiplayer bandits without collision information. On the other side, the shared randomness assumption is equivalent to the knowledge of a common seed for all queues, which then use this common seed for their random generators. A similar assumption is used in multiplayer bandits \citep{bubeck2020ccoperative}.

\algoname\ is inspired by the celebrated $\varepsilon$-greedy strategy. 
With probability $\varepsilon_t = (N+K)t^{-\frac{1}{5}}$, at each time step, queues explore the different parameters $\lambda_i$ and $\mu_i$ as described below. Otherwise with probability $1-\varepsilon_t$, they exploit the servers. Each queue $i$ then sends a packet to a server following a policy solely computed from its local estimates $\hat{\lambda}^i, \hat{\mu}^i$ of the problem parameters $\lambda$ and $\mu$. 
The shared randomness is here used so that exploration simultaneously happens for all queues. If exploration/exploitation was not synchronized between the queues, an exploiting queue could collide with an exploring queue, biasing the estimates $\hat{\lambda}^i, \hat{\mu}^i$ of the latter.
\begin{algorithm}[ht]
\DontPrintSemicolon
\SetKwInOut{Input}{input}
\Input{ $N$ (number of queues), $i \in [N]$ (queue \textit{id})}
	\For{$t=1, \ldots, \infty$}{
	$\hat{P} \gets\phi(\hat{\lambda}, \hat{\mu})$ \ and \ $\hat{A}\gets\psi(\hat{P})$ where $\phi$ and $\psi$ are resp. defined by \cref{eq:phi1,eq:birkhoff1}\;
	Draw $\omega_1 \sim \bernoulli((N+K)t^{-\frac{1}{5}})$ and $\omega_2 \sim \mathcal{U}(0,1)$ \tcp*{shared randomness}
	\lIf(\tcp*[f]{exploration}){$\omega_1=1$}{$\explore(i)$}
	\lElse(\tcp*[f]{exploitation}){Pull $\hat{A}(\omega_2)(i)$}}
	\caption{\algoname\label{algo1}}
\end{algorithm}
\paragraph{Exploration.} When exploring, queues choose either to explore the servers' parameters $\mu_k$ or the other queues' parameters $\lambda_i$ as described in \cref{alg:explore} below. 
In the former case, all queues choose different servers at random (if they have packets to send). These rounds are used to estimate the servers means: $\hat{\mu}^i_k$ is the empirical mean of server $k$ observed by the queue $i$ for such rounds. Thanks to the shared randomness, queues pull different servers here, making the estimates unbiased.

In the latter case, queues explore each other in a pairwise fashion. When queues $i$ and $j$ explore each other at round $t$, each of them sends their \textbf{most recent} packet to some server $k$, chosen uniformly at random, if and only if a packet appeared during round~$t$. 
In that case, we say that \textit{the queue $i$ explores $\lambda_j$} (and vice versa). To make sure that $i$ and $j$ are the only queues choosing the server $k$ during this step, we proceed as follows: 
\begin{itemize}[leftmargin=1cm, topsep=0pt] \setlength\itemsep{0.5em}
    \item queues sample a matching $\pi$ between queues at random. To do so, the queues use the same method to plan an all-meet-all (or round robin) tournament, for instance Berger tables \citep{Berger}, and choose uniformly at random which round of the tournament to play. If the number of queues $N$ is odd, in each round of the tournament, one queue remains alone and does nothing.
    \item the queues draw the same number $l \sim \mathcal{U}([K])$ with their shared randomness. For each pair of queues $(i,j)$ matched in $\pi$, associate $k_{(i,j)} = l + \min(i,j) \mod{K}+1$ to this pair. The queues $i$ and $j$ then send to the server $k_{(i,j)}$.
\end{itemize}

As we assumed that the server breaks ties in the packets' age uniformly at random, the queue $i$ clears with probability $(1-\frac{\lambda_j}{2})\bar{\mu}$, where $\bar{\mu} = \frac{1}{K}\sum_{k=1}^K \mu_k$. Thanks to this, $\lambda_j$ is estimated by queue $i$ as:
\begin{equation}\label{eq:estimlambda}
\textstyle    \hat{\lambda}^i_j = 2- 2\hat{S}^i_j/\tilde{\mu}^i,
\end{equation}
where $\tilde{\mu}^i = \frac{\sum_{k=1}^K N_k^i \hat{\mu}_k^i}{\sum_{k=1}^K N_k^i}$, $N_k^i$ is the number of \textit{exploration} pulls of server $k$ by queue $i$ and $\hat{S}^i_j$ is the empirical probability of clearing a packet observed by queue $i$ when exploring $\lambda_j$.
\begin{algorithm}[ht]
\DontPrintSemicolon
\SetKwInOut{Input}{input}
\Input{$i \in [N]$ \tcp*{queue \textit{id}}}
    $k \gets 0$ \;
    Draw $n \sim \mathcal{U}([N+K])$\tcp*{shared randomness}
	\uIf(\tcp*[f]{explore $\mu$}){$n\leq K$}{$k \gets n + i \mod{K} +1$ \; Pull $k$ ;\quad
	Update $N_k$ and $\hat{\mu}_k$}
	\Else(\tcp*[f]{explore $\lambda$}){

	Draw $r \sim \mathcal{U}([N])$ and $l \sim \mathcal{U}([K])$\tcp*{shared randomness}
	$j \gets r ^{\text{th}}$ opponent in the all-meet-all tournament planned according to Berger tables\;
	$k \gets l + \min(i,j) \mod{K}+1$\; 
	\If(\tcp*[f]{explore $\lambda_j$ on server $k$}){$k \neq 0$ and packet appeared at current time step}{Pull $k$ with most recent packet ;\quad
	Update $\hat{S}_j$ and $\hat{\lambda}_j$ according to \cref{eq:estimlambda}}}
	\caption{\label{alg:explore}\explore}
\end{algorithm}
\begin{remark} 
The packet manipulation when exploring $\lambda_j$ strongly relies on the servers tie breaking rules (uniformly at random). If this rule was unknown or not explicit, the algorithm can be adapted: when queue $i$ explores $\lambda_j$, queue $j$ instead sends the packet generated at time $t-1$ (if it exists), while queue $i$ still sends the packet generated at time $t$. In that case, the clearing probability for queue $i$ is exactly $(1-\lambda_j)\bar{\mu}$, allowing to estimate $\lambda_j$. Anticipating the nature of the round $t$ (exploration vs. exploitation) can be done by drawing $\omega_1 \sim \bernoulli(\varepsilon_t)$ at time $t-1$. If $\omega_1=1$, the round $t$ is exploratory and the packet generated at time $t-1$ is then kept apart by the queue $j$.
\end{remark}
To describe the exploitation phase, we need a few more notations. We denote by $\bisto$ the set of doubly stochastic matrices (non-negative matrices such that each of its rows and columns sums to $1$) and by $\mathfrak{S}_K$  the set of permutation matrices in $[K]$ (a permutation matrix will be identified with its associated permutation for the sake of cumbersomeness).

A \textbf{\mapphi} is a function $\phi:\mathbb{R}^{N}\times \mathbb{R}^K \to \bisto$ which, from~$(\lambda, \mu)$, returns a doubly stochastic matrix $P$ such that $\lambda_i< \left(P \mu\right)_i$ for any $i \in [N]$ if it exists (and the identity matrix otherwise). 

A \textbf{BvN} (Birkhoff von Neumann) \textbf{decomposition} is a function $\psi:\bisto \to \mathcal{P}(\mathfrak{S}_K)$  that associates to any doubly stochastic matrix $P$ a random variable $\psi(P)$ such that $\mathds{E}[\psi(P)] = P$; stated otherwise,  it expresses $P$ as a convex combination of permutation matrices. For convenience, we will represent this random variable as a function from $[0,1]$ (equipped with the uniform distribution) to $\mathfrak{S}_K$.

Informally speaking, those functions describe the strategies queues would follow in the centralized case: a \mapphi\ gives adequate marginals ensuring stability (since the queue $i$  clears in expectation $(P\mu)_i$ packets at each step, which is larger than $\lambda_i$  by definition), while a \mappsi\ describes the associated coupling to avoid collisions. Explicitly, the joint strategy is for each queue to draw a shared random variable $\omega_2 \sim \mathcal{U}(0,1)$ and to choose servers according to the permutation $\psi(\phi(\lambda,\mu))(\omega_2)$

\paragraph{Exploitation.} 
In a decentralized system, each queue $i$  computes a mapping $\hat{A}^i:=\psi(\phi(\hat{\lambda}^i, \hat{\mu}^i))$ solely based on its own estimates $\hat{\lambda}^i, \hat{\mu}^i$.
A shared variable $\omega_2 \in [0,1]$ is then generated uniformly at random and queue $i$ sends a packet to the server $\hat{A}^i(\omega_2)(i)$. If all queues knew exactly the parameters $\lambda, \mu$, the computed strategies $\hat{A}^i$ would be identical and they would follow the centralized policy described above.

However,  the estimates $(\hat{\lambda}^i, \hat{\mu}^i)$ are different between queues. The usual \mapphi s and \mappsi s in the literature are non-continuous. Using those, even queues with close estimates could have totally different $\hat{A}^i$, and thus collide a large number of times, which would impede the stability of the system. 
Regular enough \mapphi s and \mappsi s are required, to avoid this phenomenon. The design of $\phi$ and $\psi$ is thus crucial and appropriate  choices are given in the following   \cref{sec:phi,sec:psi}. Nonetheless, they can be used  in some black-box fashion, so we provide for the sake of completeness sufficient conditions for stability, as well as a general result depending on the properties of $\phi$ and $\psi$, in \cref{app:blackbox}.

\begin{remark}
The exploration probability $t^{-\frac{1}{5}}$ gives the smallest theoretical dependency in $\Delta$ in our bound. A trade-off between the proportion of exploration rounds and the speed of learning indeed appears in the proof of \cref{thm:stabilityintro}. Exploration rounds have to represent a small proportion of the rounds, as the queues accumulate packets when exploring. On the other hand, if queues explore more often, the regime where their number of packets decreases starts earlier. 
A general stability result depending on the choice of this probability is given by \cref{thm:stabilityblackbox} in \cref{app:blackbox}. \\
Yet in \cref{sec:simulations}, taking a probability $t^{-\frac{1}{4}}$ empirically performs better as it speeds up the exploration.
\end{remark}

\subsection{Choice of a \mapphi}\label{sec:phi}

Recall that a \mapphi\ takes as inputs $(\lambda, \mu)$ and returns, if possible, a doubly stochastic matrix $P$ such that
\begin{equation*}
    \textstyle\lambda_i < \sum_{k=1}^K P_{i,k} \mu_k \text{ for all } i \in [N].
\end{equation*}
The usual \mapphi s sort the vector $\lambda$ and $\mu$ in descending orders \citep{marshall1979inequalities}. Because of this operation, they are non-continuous and we thus need to design a regular \mapphi\ satisfying the above property. Inspired by the $\log$-barrier method, it is done by taking the minimizer of a strongly convex program as follows
\begin{equation}\label{eq:phi1}
\phi(\lambda, \mu) = \argmin_{P \in \bisto} \max_{i \in [N]} - \ln\Big(\sum_{j=1}^K P_{i,j}\mu_j - \lambda_i\Big) + \frac{1}{2K}\| P \|_2^2 .
\end{equation}

Although the objective function is non-smooth because of the $\max$ operator, it enforces fairness between queues and leads to a better regularity of the $\argmin$.

\begin{remark}
Computing $\phi$ requires solving a non-smooth strongly convex minimization problem. This cannot be computed exactly, but a good approximation can be quickly obtained using the scheme described in \cref{app:computephi}. If this approximation error is small enough, it has no impact on the stability bound of \cref{thm:stability}. It is thus ignored for simplicity, \ie we assume in the following that $\phi(\lambda, \mu)$ is exactly computed at each step.
\end{remark}

As required, $\phi$ always returns a matrix $P$ satisfying that $\lambda < P \mu$ if possible, since otherwise the objective is infinite (and in that case we assume that $\phi$ returns the identity matrix). Moreover, the objective function is $\frac{1}{K}$-strongly convex, which guarantees some regularity of the $\argmin$, namely local-Lipschitzness, leading to \cref{lemma:phi} below.

\begin{lemma}\label{lemma:phi} For any $(\lambda,\mu)$ with positive margin $\Delta$ (defined in \cref{EQ:margin}), 
if $\|(\hat{\lambda}-\lambda, \hat{\mu}-\mu)\|_{\infty} \leq c_1\Delta$, for any  $c_1 < \frac{1}{2\sqrt{e}+2}$, then
\begin{equation*}
\|\phi(\hat{\lambda}, \hat{\mu}) - \phi(\lambda, \mu) \|_{2} \leq \frac{c_2 K}{\Delta} \|(\hat{\lambda}-\lambda, \hat{\mu}-\mu) \|_{\infty},
\end{equation*}
where $c_2 = \frac{4}{(1-2c_1)/\sqrt{e}-2c_1}$.
Moreover, denoting $\hat{P}=\phi(\hat{\lambda}, \hat{\mu})$, it holds for any $i \in [N]$,
\begin{equation*}\textstyle
    \lambda_i \leq \sum_{k=1}^K \hat{P}_{i,k} \mu_k - \left(\frac{1-2c_1}{\sqrt{e}}-2 c_1 \right)\Delta.
\end{equation*}
\end{lemma}

The first property guarantees that if the queues have close estimates, they also have close doubly stochastic matrices $\hat{P}$. Moreover, the second property guarantees that any queue should clear its packets with a margin of order $\Delta$, in absence of collisions.

\begin{remark}
An alternative dominant mapping without the regularizing term in \cref{eq:phi1} can also be proposed. Yet, its local Lipschitz bound would also depend on the smallest difference between the $\lambda_i$ or the $\mu_i$, which can be arbitrarily small. If two parameters $\lambda_i$ or $\mu_i$ are equal, this choice of dominant mapping might lead to unstable policies. Using a regularization term in \cref{eq:phi1} thus avoids this problem, although a smaller dependency might be possible without this regularization term when the parameters $\lambda_i$ and $\mu_i$ are very distinct.
\end{remark}

\subsection{Choice of a \mappsicomplete}\label{sec:psi}

Given a doubly stochastic matrix $\hat{P}$, Birkhoff algorithm returns a convex combination of permutation matrices $P[j]$ such that $\hat{P} = \sum_{j} z[j] P[j]$. 
The classical version of Birkhoff algorithm is non-continuous in its inputs and it even holds for its extensions as the one proposed by \citet{dufosse2018further}. Yet it can be smartly modified as in \birkhoff, described in \cref{alg:birkhoff}, to get a regular \mappsi\ defined as follows for any $\omega \in (0,1)$:
\begin{gather} 
\psi(P)(\omega) = P[j_\omega] \label{eq:birkhoff1}\\
\textstyle\text{where } P = \sum_{j} z[j]P[j] \text{ is the decomposition returned by \birkhoff\ algorithm} \nonumber\\
\text{and } j_\omega \text{ verifies } \sum_{j\leq j_\omega} z[j] \leq \omega < \sum_{j\leq j_\omega+1} z[j]. \nonumber
\end{gather}
For a matrix $P$ in the following, its support is defined as $\supp{P}=\lbrace (i,j) \mid P_{i,j} \neq 0\rbrace$. 
\begin{algorithm}[H]
\DontPrintSemicolon
\SetKwInOut{Input}{input}
\Input{$\hat{P} \in \bisto$ (doubly stochastic matrix), $C\in \mathbb{R}^{K\times K}$ (cost matrix)}
    $j \gets 1$ \;
	\While{$\hat{P} \neq \mathbf{0}$}{
	$C_{i,k} \gets +\infty$ for all $(i,k) \not\in \supp{\hat{P}}$ \tcp*{remove edge $(i,k)$ in induced graph}
	$P[j] \gets \hungarian(C)$ \label{alg:computematching} \tcp*{matching with minimal cost w.r.t.\ $C$}
	$z[j] \gets \min_{(i,k) \in \supp{P[j]}} \hat{P}_{i,k}$ \;
	$\hat{P} \gets \hat{P} - z[j]P[j]$ \quad and \quad
	$j \gets j+1$}
	\Return $(z[j], P[j])_{j}$
	\caption{\label{alg:birkhoff}\birkhoff}
\end{algorithm}
Obviously $\mathbb{E}_{\omega\sim\mathcal{U}(0,1)}[\psi(P)(\omega)] = P$ and permutations avoid collisions between queues.
The difference with the usual Birkhoff algorithm happens at \cref{alg:computematching}. Birkhoff algorithm usually computes any perfect matching in the graph induced by the support of $\hat{P}$ at the current iteration. This is often done with the Hopcroft-Karp algorithm, while it is here done with the Hungarian algorithm with respect to some cost matrix $C$. 
Although using the Hungarian algorithm slightly increases the computational complexity of this step ($K^3$ instead of $K^{2.5}$), it ensures to output the permutation matrices $P[j]$ according to a fixed order defined below.

\begin{definition}\label{defn:inducedorder}
A cost matrix $C$ induces an order $\prec_C$ on the permutation matrices defined, for any $P,P' \in \permuset[K]$ by 
\begin{equation*}
  P \prec_C P' \quad \text{ iff } \quad
\textstyle\sum_{i,j} C_{i,j} P_{i,j} < \sum_{i,j}C_{i,j} P'_{i,j}.
\end{equation*}
\end{definition}

This order might be non-total as different permutations can have the same cost. However, if $C$ is drawn at random according to some continuous distribution, this order is total with probability $1$. The order $\prec_C$ has to be the same for all queues and is thus determined beforehand for all queues.

\begin{lemma}\label{lemma:birkhoff}
Given matrices $C \in \mathbb{R}^{K\times K}$ and $P \in \bisto$, \birkhoff\ outputs a sequence $(z[j], P[j])_j$ of length at most $K^2-K+1$, such that
\begin{gather*}\textstyle
    P = \sum_{j} z[j]P[j], \text{ where for all } j,\  z[j] > 0 \text{ and } P[j] \in \permuset[K].
\end{gather*}
Moreover if the induced order $\prec_C$ is total, $z[j]$ is the $j$-th non-zero element of the sequence $(z_l(P))_{1\leq l \leq K!}$ defined by
\begin{equation}\label{eq:birkhoff2}
z_j(P) = \min_{(i,k) \in \supp{P_j}} \bigg(P-\sum_{l=1}^{j-1}z_l(P)P_l\bigg)_{i,k}
\end{equation}
where $(P_j)_{1 \leq j \leq K!}$ is a $\prec_C$-increasing sequence of permutation matrices, \ie
    $P_j \prec_{C} P_{j+1}$ for all $j$.
\end{lemma}

\cref{lemma:birkhoff} is crucial to guarantee the regularity of $\psi$, given by \cref{lemma:psi}.

\begin{lemma}\label{lemma:psi}
Consider $\psi$ defined as in \cref{eq:birkhoff1} with a cost matrix $C$ inducing a total order $\prec_C$, then for any doubly stochastic matrices $P,P'$
\begin{equation*}
    \int_{0}^1 \one{\psi(P)(\omega)\neq\psi(P')(\omega)} \df\omega \leq 2^{2(K^2-K+1)} \|P-P'\|_\infty.
\end{equation*}
\end{lemma}

\cref{lemma:psi} indeed ensures that the probability of collision between two queues remains small when they have close estimates.
Unfortunately, the regularity constant is exponential in $K^2$, which yields a similar dependency in the stability bound of \cref{thm:stability}. The existence of a \mappsi\ with polynomial regularity constants remains unknown, even without computational considerations. The design of a better \mappsi\ is left open for future work and would directly improve the stability bounds, using the general result given by \cref{thm:stabilityblackbox} in \cref{app:blackbox}. 
The number of accumulated packets yet remain reasonably small in the experiments of \cref{sec:simulations}, suggesting that the bound given by \cref{lemma:psi} is not tight and might be improved in future work.

\subsection{Stability guarantees}

This section finally provides theoretical guarantees on the stability of the system when all queues follow \algoname. The success of \algoname\ relies on the accurate estimation of all problem parameters by the queues, given by \cref{lemma:concentration} in \cref{app:stabilityproof}.
After some time $\tau$, the queues have tight estimations of the problem parameters. Afterwards, they clear their packets with a margin of order $\Delta$, thanks to \cref{lemma:phi,lemma:psi}. This finally ensures the stability of the system, as given by \cref{thm:stability}.

\begin{theorem}\label{thm:stability}
For any $\eta>1$, the system where all queues follow \algoname, for any queue $i$ and any $r \in \mathbb{N}$, there exists a constant $C_r$ depending only on $r$ such that
\begin{equation*}
   \mathbb{E}[(Q_t^i)^r] \leq C_r KN \bigg(\frac{N^{\frac{5}{2}} K^{\frac{5}{2}} 2^{5(K^2-K+1)}}{\left(\min(1, K\bar{\mu})\munderbar{\lambda}\right)^{\frac{5}{4}}\Delta^5}\bigg)^{r}, \quad \text{for all } t \in \mathbb{N}.
\end{equation*}
As a consequence, for any $\eta>1$, this decentralized system is strongly stable.
\end{theorem}

Despite yielding an exponential dependency in $K^2$, this anytime bound leads to a first decentralized stability result when $\eta \in (1, \frac{e}{e-1})$, which closes the stability gap left by previous works.
Moreover it can be seen in the proof that the asymptotic number of packets is much smaller. It actually converges, in expectation, to the number of packets the queues would accumulate if they were following a stable centralized strategy from the beginning. As already noted by \citet{NIPS2016_430c3626} for a single queue, the number of packets first increases during the learning phase and then decreases once the queues have tight enough estimations, until reaching the same state as in the perfect knowledge centralized case. This is empirically confirmed in \cref{sec:simulations}.

\section{Simulations}\label{sec:simulations}

 \cref{fig:setting1,fig:setting2} compare on toy examples the stability of queues, when either each of them follows the no-regret strategy EXP3.P.1, or each queue follows \algoname. For practical considerations, we choose the exploration probability $\varepsilon_t=(N+K)t^{-\frac{1}{4}}$ for \algoname, as the exploration is too slow with $\varepsilon_t$ of order $t^{-\frac{1}{5}}$. 
 
 These figures illustrate the evolution of the average queue length on two different instances with $N=K=4$. The code for the experiments is available at \url{gitlab.com/f_sen/queuing_systems}.
 
 In the first instance shown in \cref{fig:setting1}, for all $i \in [N]$, $\lambda_i = (N+1)/N^2$. Moreover $\mu_1 =1$ and for all $i \geq 2$, $\mu_i = (N-1)/N^2$. Here $\eta < 2$ and no-regret strategies are known to be unstable \citep{gaitonde2020stability}. It is empirically confirmed as the number of packets in each queue diverges when they follow EXP3.P.1. Conversely, when the queues follow \algoname, after a learning phase, the queues reach equilibrium and all succeed in clearing their packets.

 In the second instance shown in \cref{fig:setting2}, for all $i \in [N]$, $\lambda_i=0.55-0.1 \cdot i$ and  $\mu_i=2.1\lambda_i$. Here $\eta>2$ and both strategies are known to be stable, which is again empirically confirmed. However, \algoname\ requires more time to learn the different parameters, suggesting that individual no-regret strategies might be better on easy instances where $\eta>2$.
\begin{figure}[ht]%
\begin{minipage}{0.49\textwidth}
    \centering
    \begin{adjustbox}{clip,trim=0cm 0cm 0cm 0cm,max width=\linewidth}\input{packets_N_4_T106.pgf}
    \end{adjustbox}
    \caption{Hard instance, $\eta<2$.}%
    \label{fig:setting1}%
\end{minipage}
\begin{minipage}{0.49\textwidth}
    \centering
   \begin{adjustbox}{clip,trim=0cm 0cm 0cm 0cm,max width=\linewidth}
   \input{packets_R_T5.pgf}
   \end{adjustbox}
    \caption{Easy instance, $\eta>2$.}%
    \label{fig:setting2}%
\end{minipage}
\end{figure}

\section{Conclusion}

In this work, we showed that minimizing a more patient version of regret was not necessarily stable when the system's slack is smaller than two and we argued that some level of cooperation was then required between learning queues to reach stability. 
We presented the first decentralized learning algorithm guaranteeing stability of any queuing system with a slack larger than $1$. Our stability bound presents an exponential dependency in the number of queues and remains open for improvement, \eg  through a better dominant mapping/BvN decomposition or a tighter analysis of ours.
The proposed algorithm relies heavily on synchronisation between the queues, which all start the game simultaneously and share a common time discretisation. In particular, the shared randomness assumption merely results from this synchronisation when the players use a common random seed. 
Stability of asynchronous queues thus remains open for future work, for which Glauber dynamics approaches used in scheduling problems might be of interest \citep[see \eg][]{shah2012randomized}.

\section*{Acknowledgements}

F. Sentenac was supported by IP PARIS' PhD Funding. E. Boursier was supported by an AMX scholarship. V. Perchet acknowledges support from the French National Research Agency
(ANR) under grant number \#ANR-19-CE23-0026 as well as the support grant, as well as from
the grant ``Investissements d’Avenir'' (LabEx Ecodec/ANR-11-LABX-0047).

\bibliographystyle{plainnat}
\bibliography{biblio}
\newpage
\appendix

\section{General version of Theorem \ref{thm:stability}}\label{app:blackbox}

\algoname\ is described for specific choices of the functions $\phi$ and $\psi$ given by \cref{sec:phi,sec:psi}. It yet uses them in a black box fashion and different functions can be used, as long as they verify some key properties. This section provides a general version of \cref{thm:stability}, when the used \mapphi\ and \mappsi\ respect the properties given by \cref{ass:phi,ass:psi}.

\begin{assumption}[regular \mapphi]\label{ass:phi}
There are constants $c_1,c_2 > 0$ and a norm $\|\cdot \|$ on $\mathbb{R}^{K\times K}$, such that if $\|(\hat{\lambda}-\lambda, \hat{\mu}-\mu) \|_{\infty}\leq c_1 \Delta$, then
\begin{equation*}
\|\phi(\hat{\lambda}, \hat{\mu}) - \phi(\lambda, \mu) \| \leq L_{\phi} \cdot \|(\hat{\lambda}-\lambda, \hat{\mu}-\mu) \|_{\infty}.
\end{equation*}

Moreover, $\hat{P}=\phi(\hat{\lambda}, \hat{\mu})$ is doubly stochastic and for any $i \in [N]$,
\begin{equation*}
  \textstyle \lambda_i \leq \sum_{k=1}^K \hat{P}_{i,k} \mu_k - c_2 \Delta.
\end{equation*}
\end{assumption}

\begin{assumption}[regular \mappsi]\label{ass:psi}
Consider the same norm $\|\cdot \|$ as \cref{ass:phi} on $\mathbb{R}^{K\times K}$. For any doubly stochastic matrices $P,P'$
\begin{gather*}
\int_{0}^1 \psi(P)(\omega) \df \omega = P\\
    \text{and } \int_{0}^1 \one{\psi(P)(\omega)\neq\psi(P')(\omega)} \df\omega \leq L_{\psi} \cdot \|P-P'\|.
\end{gather*}
\end{assumption}

\cref{lemma:phi,lemma:psi} show that the functions described in \cref{sec:phi,sec:psi} verify \cref{ass:phi,ass:psi} with the constants $L_{\phi}$ and $L_{\psi}$ respectively of order $\frac{K}{\Delta}$ and $2^{2K^2}$ with the norm $\|\cdot\|_{\infty}$. Designing a \mapphi\ and a \mappsi\ with smaller constants $L_{\phi}$ and $L_{\psi}$ is left open for future work. It would lead to a direct improvement of the stability bound, as shown by \cref{thm:stabilityblackbox}.

\begin{theorem}\label{thm:stabilityblackbox}
Assume all queues follow \algoname, using an exploration probability $\varepsilon_t = x t^{-\alpha}$ with $x>0, \alpha \in (0,1)$ and functions $\phi$ and $\psi$ verifying \cref{ass:phi,ass:psi} with the constants $L_{\phi},L_{\psi}$. The system is then strongly stable and for any $r \in \mathbb{N}$, there exists a constant $C_r$ such that:
\begin{equation*}
    \mathbb{E}[(Q_t^i)^r] \leq C_r \left( \frac{x^{r/\alpha}}{\Delta^{r/\alpha}}+KN\left( \frac{N^2K L_{\phi}^2L_{\psi}^2}{\min(1, K\bar{\mu})\munderbar{\lambda} \Delta^2 x} \right)^{\frac{r}{1-\alpha}} \right),\quad \text{for all } t \in \mathbb{N}
\end{equation*}
\end{theorem}

The proof directly follows the lines of the proof of \cref{thm:stability} in \cref{app:stabilityproof} and is thus omitted here. 
From this version, it can be directly deduced that $\alpha=\frac{1}{5}$ gives the best dependency in $\Delta$ for \algoname. Moreover the best choice for $x$ varies with $r$. When $r\to\infty$, it actually is $x= N^{\frac{2}{5}}K^{\frac{3}{5}}2^{\frac{4}{5}K^2}$ for \algoname. The choice $x=N+K$ is preferred for simplicity and still yields quite similar problem dependent bounds.

\section{Efficient computation of \texorpdfstring{$\phi$}{ϕ}} \label{app:computephi}

As mentioned in \cref{sec:phi}, computing exactly $\phi(\hat{\lambda}, \hat{\mu})$ is not possible. Even efficiently approximating it is not obvious, as the function to minimize is neither smooth nor Lipschitz. We here describe how an approximation of $\phi$ can be efficiently computed with guarantees on the approximation error. 

First define the empirical estimate of the margin $\Delta$:
\begin{equation*}
    \hat{\Delta} \coloneqq \min_{k\in[N]} \frac{1}{k}\left(\sum_{i=1}^k \hat{\mu}_{(i)} - \hat{\lambda}_{(i)} \right).
\end{equation*}
It can be computed in time $\bigO{N\log(N)}$ as it only requires to sort the vectors $\hat{\lambda}$ and $\hat{\mu}$. If $\hat{\Delta}\leq0$, then the value of the optimization problem is $+\infty$ and any matrix can be returned. 
Assume in the following $\hat{\Delta}>0$. Similarly to the proof of \cref{lemma:phi}, it can be shown that the value of the optimization problem is smaller than $-\ln(\hat{\Delta}/\sqrt{e})$. 
Noting by $\bisto$ the set of $K\times K$ doubly stochastic matrices, the optimization problem given by \cref{eq:phi1} is then equivalent to

\begin{equation}\label{eq:phi1constr}
\argmin_{P \in \mathcal{X}} g(P),
\end{equation}
where 
\begin{gather*}
\textstyle
\mathcal{X} = \Bigg\lbrace P \in \bisto\mid\  \forall i \in [N], \sum_{j=1}^K P_{i,j}\mu_j - \lambda_i \geq \frac{\hat{\Delta}}{\sqrt{e}}\Bigg\rbrace,\\
\textstyle\text{and } g(P) = \max_{i \in [N]} - \ln(\sum_{j=1}^K P_{i,j}\mu_j - \lambda_i) + \frac{1}{2K}\| P \|_2^2.
\end{gather*}

Thanks to this new constraint set, the objective function of \cref{eq:phi1constr} is now $(\frac{\sqrt{e}}{\Delta}+1)$-Lipschitz. We can now use classical results for Lipschitz strongly convex minimization to obtain convergence rates of order $\frac{1}{t}$ for the projected gradient descent algorithm \citep[see \eg][Theorem 3.9]{bubeck2014}.
These results yet assume that the projection on the constraint set can be exactly computed in a short time. This is not the case here, but it yet can be efficiently approximated using interior point methods \citep[see \eg][Section 5.3]{bubeck2014}, which has a linear convergence rate. If this approximation is good enough, similar convergence guarantees than with exact projection can be shown similarly to the original proof.

\cref{alg:computephi} then describes how to quickly estimate $\phi(\hat{\lambda}, \hat{\mu})$, where $\hat{\proj}_{\mathcal{X}}$ returns an approximation of the orthogonal projection on the set $\mathcal{X}$ and $\partial g$ is a sub-gradient of $g$. It uses an averaged value of the different iterates, as the last iterate does not have good convergence guarantees.

\begin{algorithm}[ht]
\DontPrintSemicolon
\SetKwInOut{Input}{input}
\Input{ function $g$, constraint set $\mathcal{X}$, $P^0 \in \mathcal{X}$}
	$P, \hat{P} \gets P^0$ \;
	\For{$t=1, \ldots, n$}{
	$P \gets \approxproj{P - \frac{2N}{(t+1)} \partial g(P)}{\mathcal{X}}$ \tcp*{approximated projection}
	$\hat{P} \gets \frac{t}{t+2} \hat{P} + \frac{2}{t+2} P$}
\Return{$\hat{P}$}
	\caption{\label{alg:computephi}Compute $\phi$}
\end{algorithm}

In practice, the approximation can even be computed faster by initializing $P^0$ in \cref{alg:computephi} with the solution of the previous round $t-1$.

\section{Unstable No-Policy regret system example }\label{app:counterexample}

\begin{algorithm}[ht]
\DontPrintSemicolon
\SetKwInOut{Input}{input}
\Input{ $w_k$, $N$, $\alpha$, $\lambda = (1/N,\ldots,1/N)$, $\mu= (2(N-d)/N^2,\ldots,2(N-d)/N^2)$}
\For{$k=1, \ldots, \infty$}{
	\For{$t=1, \ldots, \lceil \alpha w_k \rceil$ }{
	Queues $2i$ and $2i+1$ play server $2i+t \mod N$\tcp*{stage 1}}
	\For{$t=\lceil \alpha w_k \rceil +1, \ldots,  w_k$ }{
	Queue $i$ plays server $i+t \mod N$\tcp*{stage 2}}}
	\caption{\label{alg:example}Unstable No-policy regret system example}
\end{algorithm}

\begin{lemma}\label{lem:exunstable}
Consider the system where the queues play according to the policy described in \cref{alg:example} over successive windows of length $w_k=k^2$. If $\alpha >1-\frac{d}{N-d}$, the system is not stable.
\end{lemma}

\begin{proof} Note that the system is equivalent to a system where each queue or pair of queue would always pick the same server. For simplicity, the analysis deals with that equivalent system.   Also, wlog, we analyse the subsystem with the two first queues and the two first servers. Let $\left\{B_{t}^{i}\right\}_{i \in[n], t \geq 1}$ be the independent random variables indicating the arrival of a packet on queue $i$ at time $t$, $\left\{S_{t}^{i}\right\}_{i \in[n], t \geq 1}$ be the indicators that server $j$ would clear a packet at iteration ${\ell}$ if one were sent to it.
For each queue $i \in[N]$ and $t \geq 0$, we have by Chernoff bound
$$
\operatorname{Pr}\left(\left|\sum_{t=1}^{\ell} B_{t}^{i}-\lambda_{i} \ell\right| \geq \sqrt{\ell\ln (\ell)}\right) \leq \frac{2}{\ell^{2}}.
$$
The same holds for each queue, thus the probability that this event happens for queue $1$ or queue $2$ is at most, $\frac{4}{\ell^{2}}$. As it is summable in $\ell$, The Borel-Cantelli Lemma implies that, for large enough $\ell$, almost surely, for any $i\in[2]$:
\begin{equation}\label{eq:received_packets}
\sum_{\ell=1}^{\ell} B_{t}^{i} = \lambda_{i} \ell \pm  \tilde{\mathcal{O}}\left(\sqrt{\ell}\right). 
\end{equation}

Let $W_k = \sum_{i=1}^{k}w_i$. Note that $W_{k}=\Theta\left(k^{3}\right)=\Theta\left(w_{k}^{3 / 2}\right)$. Again by Chernoff bound and Borel-Cantelli, for large enough $k$, almost surely, for any $i\in\lbrace 1, 2\rbrace$:
\begin{equation}\label{eq:serverpackets}
\sum_{t=W_{k-1}}^{W_{k-1}+\lceil \alpha w_k \rceil } S_{t}^{i}= \mu_i \alpha w_k \pm \tilde{\mathcal{O}}\left(\sqrt{w_{k}}\right), \ \sum_{t=W_{k-1}+\lceil \alpha w_k \rceil }^{W_{k}} S_{t}^{i}= \mu_i(1- \alpha) w_k \pm \tilde{\mathcal{O}}\left(\sqrt{w_{k}}\right).
\end{equation}

Thus, for any large enough $k$, the total number of packet in both queues at time $W_k$ is almost surely lower bounded as:
\begin{align}\label{eq:totalpacketsnodeviation}
    Q_{W_k}^1+Q_{W_k}^2\geq& \sum_{t=1}^{W_k}(B_{t}^{1}+B_{t}^{2})-\sum_{t=1}^{W_k}S_{t}^{1}-\sum_{l=1}^k\left(\sum_{t=W_{l-1}+\lceil\alpha w_l \rceil}^{W_l}S_{t}^{2}\right)\\
    \geq& \left[\frac{2}{N}-\frac{2(N-d)}{N^2}-(1-\alpha)\frac{2(N-d)}{N^2}\right]W_k- \tilde{\mathcal{O}}\left(W_k^{2/3}\right)\\
    \geq & \frac{2\left[\alpha(N-d)-(N-2d)\right]}{N^2}W_k-\tilde{\mathcal{O}}\left(W_k^{2/3}\right)
\end{align}

which is a diverging function of $W_k$. Note that this result also holds for any two pair of queues $(2i-1,2i)$, with $i \in [N/2]$.

\end{proof}

\begin{lemma}\label{eq:packetsclearedinit}
Consider the same setting as in \cref{lem:exunstable}. For any $i\in [N]$, for any large enough $k$, queue $i$ clears
\begin{equation*}
    \left(\frac{N-d}{N^2}+(1-\alpha)\frac{N-d}{N^2} +o(1)\right)w_k
\end{equation*}
 packets almost surely over window $w_k$.
\end{lemma}

\begin{proof}

The proof starts by showing that for any large enough $t$, all the queues hold roughly the same number of packets. Then, as they receive roughly the same number of packets over a time window and we can compute the approximate total number of packets cleared, the results follows.

Let $T_i^t$ be the age of the oldest packet in queue $i$ at time $t$. By Chernoff bound, 
$$
\operatorname{P}(|T_i^t-NQ_i^t|\geq N\sqrt{t\ln(t)})\leq \frac{2}{t^2}.
$$
Thus, using the Borel-Cantelli lemma, for any queue $i$, almost surely, for any large enough $k$ and any $t \in [W_{k-1}+1,W_{k}]$, 
\begin{equation}\label{eq:packet_age}
|T_i^t-NQ_i^t|\leq N\sqrt{t\ln(t)}=\tilde{\mathcal{O}}(w_k^{3/4}).\\
\end{equation}

 For any $(i,j) \in[N]^2$, define $$\phi_t^+(i,j):=\left(Q^i_{t} -Q^j_{t}-2N\sqrt{t\ln(t)}\right)_+ \text{ and }\phi_t^-(i,j):=\left(Q^i_{t} -Q^j_{t}+2N\sqrt{t\ln(t)}\right)_- .$$

Let $C_t^i$ be the indicator function that queue $i$ clears a packet at iteration $t$. Note that for any large enough $t$, $\phi_t^+(i,j)$ is a supermartingale. Indeed, 
\begin{align*}
  \mathbb{E}[\phi_{t+1}^+(i,j)|\phi_{1:t}^+(i,j)] \leq &\phi_{t}^+(i,j)+  \mathbb{E}[B_{t}^{i}-B_{t}^{j}|\phi_{1:t}^+(i,j) ]-\mathbb{E}[C_{t}^{i}-C_{t}^{j}|\phi_{1:t}^+(i,j) ]\\
  \leq &\phi_{t}^+(i,j). 
\end{align*}
The second inequality comes from \cref{eq:packet_age}, that implies that for any large enough $t$, if $\phi_t^+(i,j)$ is strictly positive, queue $i$ holds the oldest packet and thus clears one with higher probability than queue $j$. 
By the same arguments, $\phi_t^-(i,j)$ a submartingale. Also, $|\phi_{t+1}^+(i,j)-\phi_{t}^+(i,j)|\leq 2(N+1)$ for any $t\geq0$, and the same holds for $\phi_t^-(i,j)$. Let $\tau_{ij}$ be the stopping time of the smallest iteration after which \cref{eq:packet_age} always holds for queues $i$ and $j$. By Azuma-Hoeffding's inequality,
$$
\operatorname{Pr}\left(\phi_{\ell}^+(i,j)-\phi_{\tau_{ij}}^+(i,j) \geq 3(N+1)\sqrt{\ell\ln (\ell)}\right) \leq \frac{2}{\ell^{2}}
$$
and
$$\operatorname{Pr}\left(\phi_{\ell}^- (i,j)-\phi_{\tau_{ij}}^+(i,j) \leq -3(N+1)\sqrt{\ell\ln (\ell)}\right) \leq \frac{2}{\ell^{2}}.
$$

This, together with a union bound and  Borel-Cantelli's Lemma implies that almost surely, for any large enough $t$, for any $(i,j) \in[N]^2$ 
\begin{equation}\label{eq:Qt}
    Q^i_{t} -Q^j_{t} = \tilde{\mathcal{O}}\left(\sqrt{t}\right).
\end{equation}
This with \cref{eq:totalpacketsnodeviation} implies that for any large enough $k$, for any $i\in [N]$, almost surely,
\begin{align*}
    Q_{W_k}^i\geq & \frac{\left[\alpha(N-d)-(N-2d)\right]}{N^2}W_k-\tilde{\mathcal{O}}\left(W_k^{2/3}\right).
\end{align*}
This means that for any large enough $k$, every queue holds at least one packet over the whole window $w_k$. This and \cref{eq:serverpackets} is already enough to show that for any time-window $w_k$, for any large enough $k$, the total number of packets cleared by any couple of queue $(2i-1,2i)$, $i \in[N/2]$ is:
\begin{equation*}\label{eq:packetsclearedtogether}
    2\left(\frac{N-d}{N^2}+(1-\alpha)\frac{N-d}{N^2}\right)w_k + \tilde{\mathcal{O}}\left(\sqrt{w_k}\right).
\end{equation*}

During time window $w_k$, according to \cref{eq:received_packets}, both every queue receives $\alpha w_k/N+\tilde{\mathcal{O}}\left(w_k^{3/4}\right)$ packets almost surely for any large enough $k$. \cref{eq:Qt} implies that for any $i \in [N/2]$
$$Q^{2i-1}_{W_k} -Q^{2i}_{W_k} = \tilde{\mathcal{O}}\left(w_k^{3/4}\right) \text{ and }Q^{2i-1}_{W_
{k-1}} -Q^{2i}_{W_{k-1}} = \tilde{\mathcal{O}}\left(w_k^{3/4}\right).$$ Therefore, over each time-window $w_k$, for any large enough $k$, each queue clears 
\begin{equation*}
    \left(\frac{N-d}{N^2}+(1-\alpha)\frac{N-d}{N^2} +o(1)\right)w_k
\end{equation*}
 packets almost surely.
\end{proof}

\begin{lemma}\label{lem:nopolicyregret}
Consider again the system where the queues play according to the policy described in \cref{alg:example} over successive windows of length $w_k=k^2$. If $\alpha<1-\frac{1}{N-1}$, the queues have $\smallo{w_k}$ policy regret in all but finitely many of the windows.
\end{lemma}

Wlog, let us consider that queue $1$ deviates, and plays at every iteration a server chosen from the probability distribution $\textbf{p}=(p_1,...,p_N)$, with $p_i$ the probability to play server $i$. To upper bound the number of packets queue $1$ clears over each time window, we can assume it always has priority over queue $2$ and ignore it in the analysis. 

 Before proving \cref{lem:nopolicyregret}, we prove the following technical one.
 
 \begin{lemma}\label{lem:packetsclearedstage1}
 Consider that a queue deviates from the strategy considered in \cref{lem:nopolicyregret} and plays at every iteration a server chosen from the probability distribution $\textbf{p}=(p_1,...,p_N)$, with $p_i$ the probability to play server $i$. For any large enough $k$, almost surely, the number of packets the deviating queue clears of the first stage of the $k^{\text{th}}$ window is
\begin{equation*}
 \left(\frac{1}{2}+\frac{1}{N}\right)\frac{2(N-d)}{N^2}\alpha w_k +\tilde{\mathcal{O}}\left(w_k^{3/4}\right).
\end{equation*}
 \end{lemma}

\begin{proof}

The proof starts by showing that for any large enough $t$, every non-deviating queue holds approximately the same number of packets.  

Fist note that for any large enough $t$, \cref{eq:packet_age} still holds surely for any queue $i$.  For any $(i,j) \in\{3,\ldots,N\}^2$, define 
$$\phi_{\ell}^+(i,j):=\left(Q^i_{\lceil \ell N\rceil} -Q^j_{\lceil \ell N\rceil}-4N\sqrt{\lceil \ell N\rceil\ln(\lceil \ell N\rceil)}\right)_+$$
and
$$\phi_{\ell}^-(i,j):=\left(Q^i_{\lceil \ell N\rceil} -Q^j_{\lceil \ell N\rceil}+4N\sqrt{\lceil \ell N\rceil\ln(\lceil \ell N\rceil)}\right)_- .$$

For any interval $[\lceil \ell N\rceil,\lceil (\ell+1) N\rceil]$ where \cref{eq:packet_age} holds for queues $1,i$ and $j$, if $\phi_{\ell}^+(i,j)$ is strictly positive, then 
$$
\mathbb{E}\left[\sum_{t=\lceil \ell N\rceil}^{\lceil (\ell+1) N\rceil}C^j_t-C^i_t\bigg\rvert\phi_{1:t}^+(i,j) \right]\leq 0.
$$
Indeed, if $\phi_{\ell}^+(i,j)$ is strictly positive and \cref{eq:packet_age} holds, queue $i$ holds the oldest packets throughout the interval. Also, queue $i$ and queue $j$ collide with queue $1$ the same number of times over the interval in expectation, and if at one iteration of the interval, queue $1$ holds an older packet than queue $i$, it holds an older packet than queue $j$ over the whole interval. Thus $\phi_{\ell}^+(i,j)$ is a submartingale. By the same arguments, $\phi_{\ell}^+(i,j)$ is a supermartingale. Also,  $|\phi_{\ell+1}^+(i,j)-\phi_{\ell}^+(i,j)|\leq 4(N+1)^2$ and the same holds for $\phi_{\ell}^-(i,j)$. Finishing with the same arguments used to prove \cref{eq:Qt},  almost surely, for any $(i,j) \in\{3,\ldots,N\}^2$, 
\begin{equation}\label{eq:Qt2}
    Q^i_{t} -Q^j_{t} = \tilde{\mathcal{O}}\left(\sqrt{t}\right).
\end{equation}

We now show that for any large enough $t$, queue $1$ can not hold many more packets than the non-deviating queues. Define 
$$\phi_t^+ := \left(Q^1_{t}- \max_{i\geq3}Q^i_{t}-2N\sqrt{t\ln(t)}\right)_+.$$
Once again, at every iteration where $\phi_t^+ $ is strictly positive and \cref{eq:packet_age} holds, queue $1$ holds the oldest packet and thus has priority on whichever server it chooses. This implies that for any large enough $t$, $\phi_t^+$ is a supermartingale. It also holds that for any $t\geq 0$, $|\phi_{t+1}^+-\phi_t^+  |\leq 2(N+1)$. Thus, with the same arguments used to prove \cref{eq:Qt}, almost surely, 
\begin{equation}\label{eq:Q1max}
    \left(Q^1_{t}- \max_{i\geq3}Q^i_{t}\right)_+ = \tilde{\mathcal{O}}\left(\sqrt{t}\right).
\end{equation}

With that at hand, we prove that for any large enough $k$, queue $1$ does not get priority often over the other queues during the first stage of the $k^{\text{th}}$ window.
For any $i \in \{2,\ldots,N/2\}$, pose:
$$
\psi_\ell^i= \frac{1}{2}\left(Q_{\lceil \ell N\rceil}^{2i-1}+Q_{\lceil \ell N\rceil}^{2i}\right)-Q_{\lceil \ell N\rceil}^1-\frac{2(N-d)}{N^3}(\lceil \ell N\rceil-W_{k-1})
$$

For any $\ell$  s.t. $\{\lceil \ell N\rceil;\lceil (\ell+1) N-1\rceil\}$ is included in the first phase of a window, we have
\begin{align*}
   \sum_{t=\lceil \ell N\rceil}^{\lceil (\ell+1) N\rceil-1}\mathbb{E}\left[C_{t}^{1}\bigg\rvert\psi_{1:\ell}^+(i,j)\right] &\geq \sum_{t=\lceil \ell N\rceil}^{\lceil (\ell+1) N\rceil-1} \mathbb{E}\left[S_i^t \mathds{1}_{\{\text{queue $1$ and only queue $1$ picks server $i$}\}}\bigg\rvert\psi_{1:\ell}^+(i,j)\right]\\
   &\geq \frac{N-d}{N}+\frac{2(N-d)}{N^2}
\end{align*}
as well as
\begin{align*}
   \sum_{t=\lceil \ell N\rceil}^{\lceil (\ell+1) N\rceil-1}\mathbb{E}\left[\frac{1}{2}\left(C_{t}^{2i}+C_{t}^{2i-1}\right)\bigg\rvert\psi_{1:\ell}^+(i,j)\right] &\leq \sum_{t=\lceil \ell N\rceil}^{\lceil (\ell+1) N\rceil-1} \mathbb{E}\left[\frac{1}{2}S_{i+t \mod N}^t \bigg\rvert\psi_{1:\ell}^+(i,j)\right]\\
   &\leq\frac{N-d}{N}.
\end{align*}
Those two inequalities imply:
\begin{align*}
    \mathbb{E}[\psi_{\ell+1}^i|\psi_{1:\ell}^+(i,j)] =& \psi_{\ell}^+(i,j) + \sum_{t=\lceil \ell N\rceil}^{\lceil (\ell+1) N\rceil-1}\mathbb{E}\left[\frac{1}{2}(B_{t}^{2i}-B_{t}^{2i-1})-B_{t}^{1}\bigg\rvert\psi_{1:\ell}^+(i,j)\right] \\
    &-\sum_{t=\lceil \ell N\rceil}^{\lceil (\ell+1) N\rceil-1}\mathbb{E}\left[\frac{1}{2}(C_{t}^{2i}-C_{t}^{2i-1})-C_{t}^{1}\bigg\rvert\psi_{1:\ell}^+(i,j)\right]- \frac{2(N-d)}{N^2}\\
    \geq&  \psi_{\ell}^+(i,j).
\end{align*}
Thus, for any $\ell$  s.t. $\{\lceil \ell N\rceil;\lceil (\ell+1) N-1\rceil\}$ is included in the first phase of a window, $\psi_{\ell}^i$ is a submartingale. Moreover, for any $\ell\geq 0$, $|\psi_{\ell+1}^i-\psi_{\ell}^i|\leq 3N$. Thus, by Azuma-Hoeffding's inequality, for any $\ell$ s.t.$ \{\lceil \ell N\rceil;\lceil (\ell+1) N-1\rceil\} \subset [W_{k-1},W_{k-1}+\alpha w_k]$,
$$
\operatorname{Pr}\left(\psi_\ell ^i-\psi_{W_k}^i \leq -6N\sqrt{ \ell N \ln(\ell N)}\right)\leq \frac{1}{(\ell N)^2}.
$$
Borel-Cantelli's lemma implies, that for any large enough $\ell$ s.t.$ \{\lceil \ell N\rceil;\lceil (\ell+1) N-1\rceil\} \subset [W_{k-1},W_{k-1}+\alpha w_k]$, almost surely:
$$
\psi_{\ell}^i \geq \psi_{W_k}^i  -6N\sqrt{ \ell N \ln(\ell N)}. 
$$
This and \cref{eq:Q1max} applied at $t=W_k$, imply that for any large enough $k$, for any $t \in [W_{k-1},W_{k-1}+\alpha w_k]$,
\begin{align*}
  \frac{1}{2}\left(Q_{t}^{2i-1}+Q_{t}^{2i}\right)\geq& Q_{\lceil \ell N\rceil}^1+\frac{2(N-d)}{N^3}(t-W_{k-1})+\psi_{W_k}^i  -\tilde{O}(\sqrt{t})  \\
   \geq&Q_{\lceil \ell N\rceil}^1+\frac{2(N-d)}{N^3}(t-W_{k-1}) -\tilde{O}(w_k^{3/4}).  \\
\end{align*}

This and \cref{eq:packet_age} imply that during the first stage of the time window, queue $1$ holds younger packets than any other queues $i\geq 3$ after at most  $\tilde{\mathcal{O}}(w_k^{3/4})$ iterations. 

By Chernoff bound and the Borel-Cantelli lemma again, for any large enough $k$, almost surely, the number of packets queue $1$ clears during the first stage of the  $k^{\text{th}}$ window on servers where it does not collide with other queues is:
\begin{equation*}
   \sum_{t=W_{k-1}+1}^{W_{k-1}+\alpha w_k}\sum_{i=1}^N S_i^t \mathds{1}_{\{\text{queue $1$ and only queue $1$ picks server $i$}\}}=(\frac{1}{2}+\frac{1}{N})\frac{2(N-d)}{N^2}\alpha w_k +\tilde{\mathcal{O}}\left(\sqrt{w_k}\right).
\end{equation*}

Since we have shown that for any large enough $k$, almost surely, queue $1$ does not have priority over the other queues after at most $\tilde{\mathcal{O}}(w_k^{3/4})$ iterations, for any large enough $k$, almost surely, the number of packets queue $1$ clears of the first stage of the $k^{\text{th}}$ window is
\begin{equation*}
 \left(\frac{1}{2}+\frac{1}{N}\right)\frac{2(N-d)}{N^2}\alpha w_k +\tilde{\mathcal{O}}\left(w_k^{3/4}\right).
\end{equation*}
\end{proof}

We are now ready to prove \cref{lem:nopolicyregret}.
\begin{proof}

By Chernoff bound and the Borel-Cantelli lemma, almost surely for any large enough $k$, the number of packets queue $1$ clears during the second stage of the window on servers where it does not collide with other queues is:

\begin{equation}\label{eq:packets2stage}
   \sum_{t=W_{k-1}+\alpha w_k}^{W_{k}-1}\sum_{i=1}^N S_i^t \mathds{1}_{\{\text{queue $1$ and only queue $1$ picks server $i$}\}}=\frac{4(N-d)}{N^3}(1-\alpha) w_k +\tilde{\mathcal{O}}\left(\sqrt{w_k}\right).
\end{equation}

Suppose that during the second stage of the window, queue $1$ never gets priority over another queue. In that case, according to \cref{eq:packets2stage}  and \cref{lem:packetsclearedstage1}, for any large enough $k$, almost surely, the total number of packets cleared by queue $1$ during the time window is 
$$
\left(\frac{\alpha}{2}+\frac{2-\alpha}{N}\right)\frac{2(N-d)}{N^2}w_k +\tilde{\mathcal{O}}(w_k^{3/4}).
$$
For any large enough $k$, if $\alpha\leq1-\frac{1}{N-1}$ this is smaller than the number of packets queue $1$ would have cleared had it not deviated, according to \cref{eq:packetsclearedinit}.

On the other hand, suppose that queue gets priority over some other queue $i$
at some iteration $\tau$ of the second stage of the window. In that case, at that iteration, queue $1$ holds the oldest packets, which, according to \cref{eq:packet_age}, implies
$$
Q_1^{\tau}>Q_i^{\tau}-\tilde{\mathcal{O}}(w_k^{3/4})
$$
During the second stage of the window, for any $i\geq3$, $\gamma_t ^i:=\left(Q_i^{t}-Q_1^{t}-2N\sqrt{t\ln(t)}\right)_+$ is a supermartingale with bounded increments for any $t$ where \cref{eq:packet_age} holds for queues $1$ and $i$. Indeed, in that case, if $\gamma_t ^i$ is strictly positive, queue $i$ holds an older packet than queue $1$, and thus, whether they collide or not, it has a higher probability to clear a packet than queue $1$. Thus, by Azuma-Hoeffding and the Borel-cantelli lemma again, for any large enough $k$, almost surely,
$$Q_i^{W_k}-Q_1^{W_k}\leq Q_i^{\tau}-Q_1^{\tau}+\tilde{\mathcal{O}}(w_k^{3/4}).$$ Thus it holds that $Q_1^{W_k}\geq Q_i^{W_k}-\tilde{\mathcal{O}}(w_k^{3/4})$ for any $i\geq2$. This and \cref{eq:Q1max} imply that all the queues clear approximately the same number of packets over those time windows for any large enough $k$ almost surely. Thus queue $1$ clears
$$
\left[(2-\alpha)(N-2)+(\alpha+\frac{4-2\alpha}{N}) \right]\frac{(N-d)}{(N-1)N^2}w_k +\tilde{\mathcal{O}}\left(w_k^{3/4}\right)
$$
packets almost surely, which again is smaller than the number of packets it would have cleared had it not deviated.

Thus, the deviating queue clears almost surely less packets by time window than it would have had it not deviated on all but finitely many of the time windows, which implies that it has no policy regret on all but finitely many of the time windows.
\end{proof}

\section{Proofs of Section~\ref{sec:algo}}\label{app:proofs}

\subsection{Proof of Lemma~\ref{lemma:phi}}

We want to show that if $\|(\hat{\lambda}-\lambda, \hat{\mu}-\mu)\|_{\infty} \leq c_1\Delta$, then
\begin{equation}\label{eq:phi1regu}
\|\phi(\hat{\lambda}, \hat{\mu}) - \phi(\lambda, \mu) \|_{2} \leq \frac{c_2 K}{\Delta} \|(\hat{\lambda}-\lambda, \hat{\mu}-\mu) \|_{\infty},
\end{equation}
with the constants $c_1, c_2$ given in \cref{lemma:phi}.

\medskip

Recall that $\phi$ is defined as
\begin{equation*}
\phi(\lambda, \mu) = \argmin_{P \in \bisto} f(P, \lambda, \mu),
\end{equation*}
where $\bisto$ is the set of $K\times K$ doubly stochastic matrices and $f$ is defined as:
\begin{equation*}
f(P, \lambda, \mu) \coloneqq \max_{i \in [N]} - \ln(\sum_{j=1}^K P_{i,j}\mu_k - \lambda_i) + \frac{1}{2K}\| P \|_2^2
\end{equation*} 

Let $P^*$ and $\hat{P}^*$ be the minimizers of $f$ with the respective parameters $(\lambda, \mu)$ and $(\hat{\lambda}, \hat{\mu})$. They are uniquely defined as $f$ is $\frac{1}{K}$ strongly convex.

As the property of \cref{lemma:phi} is symmetric, we can assume without loss of generality that $f(P^*, \lambda, \mu) \geq f(\hat{P}^*, \hat{\lambda}, \hat{\mu})$.

\medskip

Given the definition of $\Delta$, we have the bound 

\begin{equation*}
 -\ln(\Delta) + \frac{1}{2} \geq f(P^*, \lambda, \mu) \geq -\ln(\Delta).
\end{equation*}

The lower bound holds because the term in the $\ln$ is at most $\Delta$ for at least one $i$. For the upper bound, some matrix $P$ ensures that the term in the $\ln$ is at least $\Delta$ for all $i$ and $\| P \|_2^2 \leq K$.

Similarly for $\hat{P}^*$, it follows:

\begin{equation*}
 -\ln((1-2c_1)\Delta) + \frac{1}{2} \geq f(\hat{P}^*, \hat{\lambda}, \hat{\mu}) \geq -\ln((1+2c_1)\Delta).
\end{equation*}

As a consequence, it holds for any $i\in[N]$:
\begin{align*}
- \ln\left(\sum_{j=1}^K \hat{P}^*_{i,j}\hat{\mu}_j - \hat{\lambda}_i \right) & \leq  f(\hat{P}^*, \hat{\lambda}, \hat{\mu}) \\
& \leq -\ln((1-2c_1)\Delta/\sqrt{e}) \\
\sum_{j=1}^K \hat{P}^*_{i,j}\hat{\mu}_j - \hat{\lambda}_i & \geq (1-2c_1)\Delta/\sqrt{e}.
\end{align*}

\medskip

Note that for any $i\in[N]$,
\begin{equation*}
\sum_{j=1}^K \hat{P}^*_{i,j}\hat{\mu}_j - \hat{\lambda}_i \leq \sum_{j=1}^K \hat{P}^*_{i,j}\mu_j - \lambda_i + 2 \|(\hat{\lambda}-\lambda, \hat{\mu}-\mu) \|_{\infty}.
\end{equation*}

It then yields the second point of \cref{lemma:phi}:
\begin{equation*}
\sum_{j=1}^K \hat{P}^*_{i,j}\mu_j - \lambda_i \geq \left((1-2c_1)/\sqrt{e} - 2c_1 \right) \Delta
\end{equation*}

Moreover, it follows
\begin{align*}
- \ln\left(\sum_{j=1}^K \hat{P}^*_{i,j}\hat{\mu}_j - \hat{\lambda}_i \right) &\geq - \ln\left(\sum_{j=1}^K \hat{P}^*_{i,j}\mu_j - \lambda_i \right) - \ln\left(1 + \frac{ 2 \|(\hat{\lambda}-\lambda, \hat{\mu}-\mu) \|_{\infty}}{\sum_{j=1}^K \hat{P}^*_{i,j}\mu_j - \lambda_i} \right) \\
& \geq - \ln\left(\sum_{j=1}^K \hat{P}^*_{i,j}\mu_j - \lambda_i \right) - \frac{ 2 \|(\hat{\lambda}-\lambda, \hat{\mu}-\mu) \|_{\infty}}{\left((1-2c_1)/\sqrt{e} - 2c_1 \right) \Delta}
\end{align*}

Recall that for a $\frac{1}{K}$-strongly convex function $g$ of global minimum $x^*$ and any $x$:
\begin{equation*}
    \left\| x-x^* \right\|_2 \leq 2K(g(x)- g(x^*))
\end{equation*}

As a consequence, it follows:

\begin{align*}
f(\hat{P}^*, \hat{\lambda}, \hat{\mu}) & \geq f(\hat{P}^*, \lambda, \mu) - \frac{ 2 \|(\hat{\lambda}-\lambda, \hat{\mu}-\mu) \|_{\infty}}{\left((1-2c_1)/\sqrt{e} - 2c_1 \right) \Delta} \\
& \geq f(P^*, \lambda, \mu) - \frac{ 2 \|(\hat{\lambda}-\lambda, \hat{\mu}-\mu) \|_{\infty}}{\left((1-2c_1)/\sqrt{e} - 2c_1 \right) \Delta} + \frac{1}{2K} \|P^* - \hat{P}^*\|_2. \\
\end{align*}

\cref{eq:phi1regu} then follows.

%
%
%
%

\subsection{Proof of Lemma~\ref{lemma:birkhoff}}

The coefficient $C_{i,j}$ is replaced by $+\infty$ as soon as the whole weight $P_{i,j}$ is exhausted. Thanks to this, the \hungarian\ algorithm does return a perfect matching with respect to the bipartite graph with edges $(i,j)$ where there remains some weight for $P_{i,j}$.
Because of this, it can be shown following the usual proof of Birkhoff algorithm \citep{birkhoff1946} that the sequence $(z[j], P[j])$ is indeed of length at most $K^2-K+1$ and is a valid decomposition of $P$.

\bigskip

Now assume that $\prec_C$ is a total order. At each iteration $j$ of \hungarian\ algorithm, denote $\tilde{P}^j = P - \sum_{s=1}^{j-1} z[s]P[s]$ the remaining weights to attribute.

Let $l_j$ be such that $P[j] = P_{l_j}$ for any iteration $j$ of \hungarian\ algorithm.

It can now be shown by induction that
\begin{equation*}
    \tilde{P}^j = P - \sum_{l=1}^{l_j}z_l(P)P_l.
\end{equation*}
where $z_l(P)$ are defined by \cref{eq:birkhoff2}. Indeed, by definition
\begin{align*}
    \tilde{P}^{j+1} & = \tilde{P}^{j} - z[j+1] P[j+1] \\
    & = \tilde{P}^{j} - z[j+1] P_{l_{j+1}}
\end{align*}

The \hungarian\ algorithm returns the minimal cost matching with respect to the modified cost matrix $C$ where the coefficients $i,k$ such that $\tilde{P}^j_{i,k}=0$ are replaced by $+\infty$. Thanks to this, $P_{l_{j+1}}$ is the minimal cost permutation matrix $P_l$ (for $\prec_C$) such that $\tilde{P}^j_{i,k} > 0$ for all $(i,k) \in \supp{P_l}$.

This means that for any $l< l_{j+1}$
\begin{equation*}
   \min_{(i,k) \in \supp{P_l}} (\tilde{P}^{j})_{i,k} = 0.
\end{equation*}
Using the induction hypothesis, this implies that $z_l(P) = 0$ for any $l_j <l< l_{j+1}$. And finally, this also implies that $z_{l_{j+1}}(P) = z[j+1]$.

\bigskip

This finally concludes the proof as $\tilde{P}^j = 0$ after the last iteration.

\subsection{Proof of Lemma~\ref{lemma:psi}}

For $z$ and $z'$ the respective decompositions of $P$ and $P'$ defined in \cref{lemma:birkhoff}, then
\begin{align*}
     \int_{0}^1 \one{\psi(P)(\omega)\neq\psi(P')(\omega)} \df\omega =  \mathbb{P}_{\omega\sim U(0,1)}\left(\psi(P)(\omega)\neq\psi(P')(\omega)\right).
\end{align*}

In the following, note $A = \psi(P)$ and $A' = \psi(P')$. It follows for $P_n$ defined as in \cref{lemma:birkhoff} 

\begin{align*}
     \int_{0}^1 \one{\psi(P)(\omega)\neq\psi(P')(\omega)} \df\omega & =  \sum_{n=1}^{K!} \mathbb{P}(A=P_n \text{ and } A' \neq P_n) \\
     & =  \frac{1}{2}\sum_{n=1}^{K!} \mathbb{P}(A=P_n \text{ and } A' \neq P_n) + \frac{1}{2}\sum_{n=1}^{K!} \mathbb{P}(A'=P_n \text{ and } A \neq P_n) \\
     & = \frac{1}{2}\sum_{n=1}^{K!} \vol{\left[\sum_{j=1}^{n-1} z_j(P), \sum_{j=1}^{n} z_j(P)\right]  \ominus \left[\sum_{j=1}^{n-1} z_j(P'), \sum_{j=1}^{n} z_j(P')\right]},
\end{align*}

where $\mathrm{vol}$ denotes the volume of a set and $A\ominus B = (A\setminus B) \cup (B\setminus A)$ is the symmetric difference of $A$ and $B$. The last equality comes from the expression of $\psi$ with respect to the coefficients $z_j(P)$, thanks to \cref{lemma:birkhoff}.

It is easy to show that $$\vol{[a, b]\ominus[c,d]} \leq \left( |c-a| + |d-b| \right) \one{b > a \text{ or } c>d}.$$ The previous equality then leads to

\begin{align}
     \int_{0}^1 \one{\psi(P)(\omega)\neq\psi(P')(\omega)} \df\omega
     & \leq \frac{1}{2}\sum_{n=1}^{K!}\left( \left|\sum_{j=1}^{n-1} z_j(P)-z_j(P') \right| + \left|\sum_{j=1}^{n} z_j(P)-z_j(P') \right| \right) \one{z_n(P)+z_n(P') > 0} \nonumber \\
     & \leq \sum_{n=1}^{K!} \left|\sum_{j=1}^{n} z_j(P)-z_j(P') \right|\one{z_n(P)+z_n(P') > 0}.\label{eq:psierror}
\end{align}
The last inequality holds because $\sum_{j=1}^{k} z_j(P)-z_j(P')$ is counted twice when $z_k(P)+z_k(P')$ is positive: when $n=k$ and for the next $n$ such that the elements are counted in the sum.

\medskip

Thanks to \cref{lemma:birkhoff}, only $2(K^2-K+1)$ elements $z_j(P)$ and $z_j(P')$ are non-zero.
Let $k_n$ be the index of the $n$-th non-zero element of $(z_s(P)+z_s(P'))_{1\leq s \leq K!}$. Note that $z_s(P')$ can be non-zero while $z_s(P)$ is zero (or conversely). 
Let also
\begin{gather*}
(i_{k_n}, j_{k_n}) \in \argmin_{(i,j) \in \supp{P_{k_n}}} P_{i,j}-\sum_{l<k_n}z_l(P)\one{(i,j) \in \supp{P_{k_n}}},\\
(i'_{k_n}, j'_{k_n}) \in \argmin_{(i,j) \in \supp{P_{k_n}}} P'_{i,j}-\sum_{l<k_n}z_l(P')\one{(i,j) \in \supp{P_{k_n}}}.
\end{gather*}

\bigskip

It then comes, thanks to \cref{lemma:birkhoff}
\begin{align*}
z_{k_n}(P)-z_{k_n}(P') & \leq P_{i'_{k_n},j'_{k_n}}-P'_{i'_{k_n},j'{k_n}}-\sum_{l<k_n}(z_l(P)-z_l(P'))\one{(i'_{k_n},j'_{k_n}) \in \supp{P_{k_n}}} \\
& \leq P_{i'_{k_n},j'_{k_n}}-P'_{i'_{k_n},j'_{k_n}}-\sum_{l<n}(z_{k_l}(P)-z_{k_l}(P'))\one{(i'_{k_n},j'_{k_n}) \in \supp{P_{k_n}}}
\end{align*}
The second inequality holds, because for $l' \not\in \lbrace k_l \mid l < 2K^2\rbrace$, the term in the sum is zero by definition of the sequence $k_l$.

A similar inequality holds for $z_{k_n}(P')-z_{k_n}(P)$, which leads to
\begin{equation*}
  |z_{k_n}(P)-z_{k_n}(P')| \leq \|P-P'\|_{\infty} + \sum_{l<n} |z_{k_l}(P)-z_{k_l}(P')|.
\end{equation*}

By induction, it thus holds
\begin{equation*}
  |z_{k_n}(P)-z_{k_n}(P')| \leq 2^{n-1} \|P-P'\|_{\infty}.
\end{equation*}

We finally conclude using \cref{eq:psierror}
\begin{align*}
 \int_{0}^1 \one{\psi(P)(\omega)\neq\psi(P')(\omega)} \df\omega & \leq \sum_{n=1}^{K!} \left|\sum_{j=1}^{n} z_j(P)-z_j(P') \right|\one{z_n(P)+z_n(P') > 0} \\
 & \leq \sum_{n=1}^{2(K^2-K+1)-1} \left|\sum_{j=1}^{k_n} z_j(P)-z_j(P') \right|\\
 & \leq \sum_{n=1}^{2(K^2-K+1)-1} \left|\sum_{l=1}^{n} z_{k_l}(P)-z_{k_l}(P') \right| \\
 & \leq \sum_{n=1}^{2(K^2-K+1)-1} \sum_{j=1}^n 2^{j-1} \|P-P' \|_{\infty}  \\
    & \leq 2^{2(K^2-K+1)}\|P-P'\|_{\infty}.
\end{align*}

In the fourth inequality, the $2(K^2-K+1)$-th term of the sum is ignored. It is indeed $0$ as $z$ and $z'$ both sum to $1$.

\subsection{Proof of Theorem~\ref{thm:stability}}\label{app:stabilityproof}

First recall below a useful version of Chernoff bound.
\begin{lemma}\label{lemma:chernoff}
For any independent variables $X_1 , \ldots , X_n$ in $[0, 1]$ and $\delta \in (0, 1)$,
\begin{equation*}
    \mathbb{P}\left(\sum_{i=1}^n X_i \leq (1-\delta)\sum_{i=1}^n \mathbb{E}[X_i] \right) \leq e^{-\frac{\delta^2 \sum_{i=1}^n \mathbb{E}[X_i]}{2}}.
\end{equation*}
\end{lemma}

We now prove the following concentration lemma.

\begin{lemma}\label{lemma:concentration}
For any time $t\geq (N+K)^5$ and $\varepsilon\in(0,\frac{1}{4})$,
\begin{gather*}
\mathbb{P}\left( |\hat{\mu}^i_k(t) - \mu_k| \geq \varepsilon \right) \leq  3 \exp\left(- \lambda_i\left(t^{\frac{4}{5}} -1\right) \varepsilon^2 \right)\\
\mathbb{P}\left( |\hat{\lambda}^i_j(t) - \lambda_j| \geq \varepsilon \right) \leq6\exp\left(-\lambda_i K  \bar{\mu}\frac{t^{\frac{4}{5}} - 1}{145} \varepsilon^2 \right).
\end{gather*}
\end{lemma}
\begin{proof}\hfill \\
\textbf{Concentration for $\hat{\mu}$.}  Consider agent $i$ in the following and denote by $N_k(t)$ the number of \textit{exploratory pulls} of this agent on server $k$ at time $t$. By definition, the probability to proceed to an exploratory pull on the server $k$ at round $t$ is at least $\lambda_i \min(t^{-\frac{1}{5}},\frac{1}{N+K})$. The term $\lambda_i$ here appears as a pull is guaranteed if a packet appeared at the current time step. Yet the number of exploratory pulls might be much larger in practice as queues should accumulate a large number of uncleared packets at the beginning.

For $t\geq (N+K)^5$, it holds:
\begin{align*}
    \sum_{n=1}^t \min(n^{-\frac{1}{5}},\frac{1}{N+K})& =  \sum_{n=1}^{(N+K)^5} \frac{1}{N+K} + \sum_{n=(N+K)^5+1}^t n^{-\frac{1}{5}}\\
    & \geq (N+K)^4 + \int_{(N+K)^5}^{t} x^{-\frac{1}{5}}\df x -1 \\
    & \geq \frac{1}{4} \left(5 t^{\frac{4}{5}} - (N+K)^4 - 4 \right) \\
    & \geq t^{\frac{4}{5}}-1.
\end{align*}

\cref{lemma:chernoff} then gives for $N_k(t)$:
\begin{align*}
    \mathbb{P}\left(N_k(t) \leq (1-\delta)\mathbb{E}[N_k(t)]\right) & \leq \exp\left(-\frac{\delta^2 \mathbb{E}[N_k(t)]}{2}\right) \\
    \mathbb{P}\left(N_k(t) \leq (1-\delta)\lambda_i\left(t^{\frac{4}{5}} - 1 \right) \right) & \leq \exp\left(-\frac{\lambda_i\delta^2 \left(t^{\frac{4}{5}} - 1 \right)}{2}\right).
\end{align*}

Which leads for $\delta=\frac{1}{2}$ to
\begin{equation}\label{eq:Njconcentration}
    \mathbb{P}\left(N_k(t) \leq \frac{\lambda_i}{2}\left(t^{\frac{4}{5}} - 1 \right) \right) \leq \exp\left(-\lambda_i\frac{ t^{\frac{4}{5}} - 1}{8}\right).
\end{equation}

\medskip

The number of exploratory pulls and the observations on the server $k$ are independent. Thanks to this, Hoeffding's inequality can be directly used as follows
\begin{equation*}
    \mathbb{P}\left( |\hat{\mu}_k^i(t) - \mu_k|  \geq \varepsilon \mid N_k(t) \right) \leq 2 \exp\left(-2 N_k(t) \varepsilon^2 \right).
\end{equation*}
Using \cref{eq:Njconcentration} now gives the first concentration inequality for $\varepsilon\leq \frac{1}{4} \leq \frac{1}{2\sqrt{2}}$:
\begin{align*}
    \mathbb{P}\left( |\hat{\mu}_k^i(t) - \mu_k|  \geq \varepsilon \right) & \leq  2 \exp\left(- \lambda_i \left(t^{\frac{4}{5}} - 1\right) \varepsilon^2 \right) +  \exp\left(-\lambda_i\frac{ t^{\frac{4}{5}} - 1}{8}\right) \\
    & \leq  3 \exp\left(- \lambda_i \left(t^{\frac{4}{5}} - 1\right) \varepsilon^2 \right).
\end{align*}

\bigskip

\textbf{Concentration for $\hat{\lambda}$.} 
Consider agent $i$ in the following. 
First show a concentration inequality for $\tilde{\mu}$. Denote by $N(t)$ the total number of exploratory pulls on servers proceeded by player $i$ at round $t$, \ie $N(t) = \sum_{k=1}^K N_k(t)$. Similarly to \cref{eq:Njconcentration}, it can be shown that 
\begin{equation*}
     \mathbb{P}\left(N(t) \leq \lambda_i K\frac{t^{\frac{4}{5}} - 1}{2} \right) \leq \exp\left(-\lambda_i K \frac{ t^{\frac{4}{5}} - 1}{8}\right).
\end{equation*}

\cref{lemma:chernoff} then gives for $\delta \in (0,1)$:
\begin{align*}
 \mathbb{P}\left( |\tilde{\mu}-\bar{\mu}|  \geq \delta \bar{\mu} \right) & \leq 2\exp\left(-\lambda_i K\delta^2 \bar{\mu}\frac{t^{\frac{4}{5}} - 1}{8} \right) + \exp\left(-\lambda_i K \frac{ t^{\frac{4}{5}} - 1}{8}\right) \\
 &  \leq 3\exp\left(-\lambda_i K\delta^2 \bar{\mu}\frac{t^{\frac{4}{5}} - 1}{8} \right).
\end{align*}

Note that $|\tilde{\mu}-\bar{\mu}| \leq \delta \bar{\mu}$ implies $|\frac{1}{\bar{\mu}}-\frac{1}{\tilde{\mu}}| \leq \frac{\delta}{(1-\delta)\bar{\mu}}$. So this gives the following inequality:

\begin{equation}\label{eq:inversemu}
 \mathbb{P}\left( \left|\frac{1}{\bar{\mu}}-\frac{1}{\tilde{\mu}}\right| \geq \frac{\delta}{(1-\delta)\bar{\mu}}\right) \leq 3\exp\left(-\lambda_i K\delta^2 \bar{\mu}\frac{t^{\frac{4}{5}} - 1}{8} \right).
\end{equation}
\bigskip

A concentration bound on $\hat{S}_j^i$ can be shown similarly for any $\delta\in(0,1)$
\begin{equation}\label{eq:concentrationSj}
    \mathbb{P}\left( \left|\hat{S}_j^i(t) - (1-\frac{\lambda_j}{2})\bar{\mu}\right|  \geq \delta\bar{\mu}\right)  \leq  3\exp\left(-\lambda_i K\delta^2 \bar{\mu}\frac{t^{\frac{4}{5}} - 1}{8} \right).
\end{equation}

Now recall that the estimate of $\lambda_j$ is defined by $\hat{\lambda}_j = 2 - \frac{2 \hat{S_j^i}}{\tilde{\mu}}$. We then have the following identity:
\begin{equation*}
    \hat{\lambda}_j - \lambda_j = 2(\frac{1}{\tilde{\mu}}-\frac{1}{\bar{\mu}}) \hat{S}_j^i + \frac{2}{\bar{\mu}}\left((1-\frac{\lambda_j}{2})\bar{\mu} - \hat{S}_j^i \right).
\end{equation*}
Since $\hat{S}_j \in [0,1]$, it yields for $\varepsilon\leq \frac{1}{4}$ and $x\in(0,1)$:
\begin{align*}
    \mathbb{P}\left( \left|\hat{\lambda}_j^i(t) - \lambda_j \right| \geq \varepsilon \right)& \leq \mathbb{P}\left( \left|2(\frac{1}{\tilde{\mu}}-\frac{1}{\bar{\mu}}) \hat{S}_j^i\right| \geq x\varepsilon \text{ or } \left|\frac{2}{\bar{\mu}}\left((1-\frac{\lambda_j}{2})\bar{\mu} - \hat{S}_j \right) \right| \geq (1-x)\varepsilon \right)  \\
    & \leq \mathbb{P}\left( \left| \frac{1}{\tilde{\mu}}-\frac{1}{\bar{\mu}} \right| \geq \frac{x\varepsilon}{2\hat{S}_j^i} \mid \hat{S}_j^i \leq (1+\frac{1-x}{8})\bar{\mu} \right) + \mathbb{P}\left(\left|\hat{S}_j^i(t) - (1-\frac{\lambda_j}{2})\bar{\mu}\right| \geq \frac{(1-x)\bar{\mu}\varepsilon}{2} \right) \\
      & \leq \mathbb{P}\left( \left| \frac{1}{\tilde{\mu}}-\frac{1}{\bar{\mu}} \right| \geq \frac{\delta}{(1-\delta)\bar{\mu}} \text{ for } \delta = \frac{4x\varepsilon}{9} \right) + \mathbb{P}\left(\left|\hat{S}_j^i(t) - (1-\frac{\lambda_j}{2})\bar{\mu}\right| \geq \frac{(1-x)\bar{\mu}\varepsilon}{2} \right)
\end{align*}

Taking $x=\frac{9}{17}$ leads to $\frac{4x}{9}=\frac{1-x}{2}$ and thus, using \cref{eq:inversemu,eq:concentrationSj}:

\begin{align*}
    \mathbb{P}\left( \left|\hat{\lambda}_j^i(t) - \lambda_j \right| \geq \varepsilon \right) &\leq 6\exp\left(-\lambda_i K \left( \frac{8}{17} \right)^2 \bar{\mu}\frac{t^{\frac{4}{5}} - 1}{32} \varepsilon^2 \right)\\
    & \leq 6\exp\left(-\lambda_i K  \bar{\mu}\frac{t^{\frac{4}{5}} - 1}{145} \varepsilon^2 \right)
\end{align*}
\end{proof}

In the following, let $c_1=0.1$ and $c_2= \frac{4}{(1-2c_1)/\sqrt{e}-2c_1} \approx 14$. For a problem instance, let the good event $\gE$ at time $t$ be defined as

\begin{equation*}
    \gE \coloneqq \left\lbrace \|(\hat{\lambda}^i - \lambda, \hat{\mu}^i - \mu) \|_{\infty} \leq \frac{0.1 \Delta^2}{2c_2 2^{2K^2}KN},\ \forall i \in [N] \right\rbrace.
\end{equation*}

As $\Delta$ is smaller than $1$, the right hand term in the definition of $\gE$ is smaller than $c_1 \Delta$. Thanks to \cref{lemma:phi,lemma:psi}, $\gE$ then guarantees that any player will collide with another player with probability at most $0.1 \Delta$, \ie, $\forall i \in [N]$,
\begin{equation*}
    \mathbb{P}_{\omega \sim \mathcal{U}(0,1)}\left( \exists j \in [N], \hat{A}_t^i(\omega) \neq \hat{A}_t^j(\omega)  \mid \gE \right) \leq 0.1\Delta.
\end{equation*}

Moreover, thanks to \cref{lemma:phi}, under $\gE$,
\begin{equation*}
   \lambda_i \leq \sum_{k=1}^K \hat{P}_{i,k} \mu_k - \left(\frac{1-2c_1}{\sqrt{e}}-2 c_1 \right)\Delta .
\end{equation*}

These two last inequalities lead to the following lemma.

\begin{lemma}\label{lemma:clear}
For $t \geq \frac{2^5 K^5}{0.08^5\Delta^5}$, denote by $\mathcal{H}_t$ the history of observations up to round $t$. Then
\begin{equation*}
    \mathbb{E}\left[ S_t^i \mid \gE, \mathcal{H}_t\right] \geq \lambda_i + 0.1 \Delta.
\end{equation*}
\end{lemma}

\begin{proof}
This is a direct consequence of the following decomposition:
\begin{align*}
    \mathbb{E}\left[ S_t^i \mid \gE, \mathcal{H}_t\right] & \geq  \overbrace{(1-(N+K)t^{-\frac{1}{5}})}^{\text{proba to exploit}}\left(\overbrace{\hat{P}_{i,k} \mu_k}^{\text{proba to clear}} - \overbrace{\mathbb{P}\left( \exists j \in [N], \hat{A}_t^i(\omega) \neq \hat{A}_t^j(\omega) \exists  \mid \gE \right)}^{\text{collision proba}}\right)\\
    & \geq (1-(N+K)t^{-\frac{1}{5}})(\lambda_i + \left(\frac{1-2c_1}{\sqrt{e}}-2 c_1 \right)\Delta  - 0.1 \Delta) \\
    & \geq (1-(N+K)t^{-\frac{1}{5}})(\lambda_i + 0.18 \Delta).
\end{align*}
The last inequality is given by $c_1=0.1$ and it leads to
\begin{equation*}
     \mathbb{E}\left[ S_t^i \mid \gE, \mathcal{H}_t\right] \geq \lambda_i + 0.18 \Delta -(N+K)t^{-\frac{1}{5}}.
\end{equation*}
For $t \geq \frac{2^5 K^5}{0.08^5\Delta^5}$, the last term is smaller than $0.08\Delta$, giving \cref{lemma:clear}.
\end{proof}

Define the stopping time
\begin{equation}\label{eq:tau}
    \tau \coloneqq \min\left\lbrace t \geq \frac{2^5 K^5}{0.08^5\Delta^5} \mid \forall t \geq s, \gE[s] \text{ holds} \right\rbrace.
\end{equation}

\begin{lemma}\label{lemma:tau}
For any integer $r\geq1$,
\begin{equation*}
    \mathbb{E}[\tau^r] =\bigO{KN \left(\frac{N^{\frac{5}{2}}K^{\frac{5}{2}} 2^{5K^2}}{\left(\min(1, K\bar{\mu})\munderbar{\lambda}\right)^{\frac{5}{4}}\Delta^5}\right)^{r}},
\end{equation*}
where the $\mathcal{O}$ notation hides constant factors that only depend on $r$.
\end{lemma}

\begin{proof}
Define for this proof $t_0 = \left\lceil\frac{2^5 K^5}{0.08^5\Delta^5}\right\rceil$. By definition, if  $\gE[t]$ does not hold for $t>t_0$, then $\tau \geq t$.
As a consequence, for any $t>t_0$ and thanks to \cref{lemma:concentration}:
\begin{align*}
    \mathbb{P}(\tau\geq t) & \leq \mathbb{P}(\neg \gE) \\
    & \leq (3eKN+6eN^2) \exp\left( -ct^{\frac{4}{5}}\right),
\end{align*}
where $c = c_0 \frac{\min(1, K \bar{\mu})\munderbar{\lambda}\Delta^4}{N^2 K^2 2^{4K^2}}$ for some universal constant $c_0 \leq 1$. 

We can now bound the moments of $\tau$:
\begin{align*}
    \mathbb{E}[\tau^r] & = r\int_0^{\infty} t^{r-1} \mathbb{P}\left(\tau \geq t \right) \df t\\
    & \leq t_0^r + (3eKN+6eN^2) r \int_{0}^{\infty} t^{r-1} e^{ -ct^{\frac{4}{5}}} \df t .
\end{align*}

Using the change of variable $u = ct^{\frac{4}{5}}$, it can be shown that $$ \int_{0}^\infty t^{r-1} e^{ -ct^{\frac{4}{5}}}\df t = \frac{5}{4} c^{-\frac{5r}{4}} \Gamma\left(\frac{5r}{4}\right),$$
where $\Gamma$ denotes the Gamma function.
It finally allows to conclude:
\begin{align*}
    \mathbb{E}[\tau^r] & = \bigO{\frac{K^{5r}}{\Delta^{5r}} + KN c^{-\frac{5r}{4}}} \\
    & = \bigO{KN \left(\frac{N^{\frac{5}{2}}K^{\frac{5}{2}} 2^{5K^2}}{\left(\min(1, K\bar{\mu})\munderbar{\lambda}\right)^{\frac{5}{4}}\Delta^5}\right)^{r}}.
\end{align*}
\end{proof}

Let $X_t$ be a random walk biased towards $0$ with the following transition probabilities:
\begin{equation}\label{eq:randomwalk}
\begin{gathered}
\mathbb{P}(X_{t+1}=X_{t}+1)=p,  \ \mathbb{P}(X_{t+1}=X_{t}-1|X_t>0)=q, \\ \mathbb{P}(X_{t+1}=X_{t}|X_t>0)=1-p-q,\ \mathbb{P}(X_{t+1}=X_{t}|X_t=0)=1-p,
\end{gathered}
\end{equation}
and $X_0=0$.

\begin{lemma}
The non-asymptotic moments of the random walk defined by \cref{eq:randomwalk} are bounded. For any $t>0,r>0$:
$$
\mathbb{E}\left[(X_t)^r\right] \leq \frac{r!}{\left(\ln\left(q/p\right)\right)^r}.
$$
\end{lemma}

\textit{Proof}: Let $\pi$ be the stationary distribution of the random walk. It verifies the following system of equations:
\begin{equation*}
    \begin{cases}
      \pi(z)=p\pi(z-1)+q\pi(z+1)+(1-p-q)\pi(z), \ \forall z>0\\
      \pi(0)=(1-p)\pi(0)+q\pi(1)\\
     \sum \pi(z)=1
    \end{cases}      
\end{equation*}
which gives:
$$
\pi(z) = \frac{q-p}{q}\left(\frac{p}{q}\right)^z.
$$
Equivalently, $\pi(z)=\Pb(\lfloor Y\rfloor=z)$ with Y an exponential random variable of parameter $\ln(q/p)$. This gives:
$$
\mathbb{E}_{X \sim \pi}\left[(X)^r\right] \leq \frac{r!}{\left(\ln\left(q/p\right)\right)^r}.
$$

Let $\tilde{X}_t$ be the random walk with the same transition probabilities as $X_t$ and $\tilde{X}_0 \sim \pi$. For any $t>0$, $\tilde{X}_t \sim \pi$. Moreover, for any $t>0$,  $\tilde{X}_t$ stochasticaly dominates $X_t$, which terminates the proof. $\hfill \Box$

\begin{proof}[Proof of \cref{thm:stability}]
For $\tau$ the stopping time defined by \cref{eq:tau}, \cref{lemma:tau} bounds its moments as follows

\begin{equation*}
    \mathbb{E}[\tau^r] = \bigO{KN \left(\frac{N^{\frac{5}{2}}K^{\frac{5}{2}} 2^{5K^2}}{\left(\min(1, K\bar{\mu})\munderbar{\lambda}\right)^{\frac{5}{4}}\Delta^5}\right)^{r}}.
\end{equation*}

Let $$p_i=\lambda_i(1-\lambda_i-0.1\Delta) \text{ and }q_i= (\lambda_i+0.1\Delta)(1-\lambda_i).$$ Let $X_t^i$ be the random walk biased towards $0$ with parameters $p_i$ and $q_i$, with $X_t^i=0$ for any $t\leq 0$. According to Lemma \ref{lemma:clear}, past time $\tau$, $Q_t^i$ is stochastically dominated by the random process $\tau+X_{t-\tau}^i$. Thus, for any $t>0$, for any $r>0$
\begin{align*}
    \mathbb{E}[\left(Q_i^t\right)^r]& \leq \max(1,2^{r-1})\left(\mathbb{E}[\tau^r]+\mathbb{E}[(X_{t-\tau} ^i)^r]\right)\\
    & = \bigO{KN \left(\frac{N^{\frac{5}{2}}K^{\frac{5}{2}}2^{5K^2}}{\left(\min(1, K\bar{\mu})\munderbar{\lambda}\right)^{\frac{5}{4}}\Delta^5}\right)^{r} + \frac{1}{\ln(q_i/p_i)^r}} \\
    &= \bigO{KN \left(\frac{N^{\frac{5}{2}}K^{\frac{5}{2}} 2^{5K^2}}{\left(\min(1, K\bar{\mu})\munderbar{\lambda}\right)^{\frac{5}{4}}\Delta^5}\right)^{r} + \Delta^{-r}} \\
    &= \bigO{KN \left(\frac{N^{\frac{5}{2}}K^{\frac{5}{2}} 2^{5K^2}}{\left(\min(1, K\bar{\mu})\munderbar{\lambda}\right)^{\frac{5}{4}}\Delta^5}\right)^{r} }.
\end{align*}
\end{proof} 

\end{document}